\documentclass[10pt, journal, twoside]{IEEEtran}

\usepackage{amsmath}
\usepackage{graphicx}
\usepackage{amsthm}
\usepackage{amssymb}
\usepackage[ruled]{algorithm2e}
\usepackage{algpseudocode}
\usepackage{microtype}
\usepackage[hidelinks]{hyperref}
\usepackage{xcolor}
\usepackage{adjustbox}
\usepackage{booktabs}
\usepackage{xparse}
\usepackage{pifont}

\usepackage[caption=false]{subfig}

\DontPrintSemicolon
\SetKwRepeat{doparallel}{do in parallel \ }{while}

\newtheorem{definition}{Definition}

\newcommand{\norm}[1]{\Vert #1 \Vert}

\newcommand{\mc}[1]{\mathcal{#1}}

\newcommand{\ti}[1]{{\tilde{#1}}}

\DeclareMathOperator*{\argmin}{argmin}
\def\*#1{\mathbf{#1}}
\def\+#1{\mathcal{#1}}
\def\<#1{\mathbb{#1}}
\newcommand{\cmark}{\ding{51}}
\newcommand{\xmark}{\ding{55}}

\usepackage{adjustbox}
\usepackage{array}

\newcolumntype{R}[2]{%
    >{\adjustbox{angle=#1,lap=\width-(#2)}\bgroup}%
    l%
    <{\egroup}%
}

\begin{document}
\title{Distributed Optimization Methods for Multi-Robot Systems: Part I --- A Tutorial}

\author{~\IEEEmembership{}Ola~Shorinwa$^1$,~Trevor~Halsted$^1$,~\IEEEmembership{}Javier~Yu$^2$,~\IEEEmembership{}Mac~Schwager$^2$~\IEEEmembership{}%
\thanks{*This project was funded in part by NSF NRI awards 1830402 and 1925030.  The second author was supported on an NDSEG Fellowship, and the third author was supported on an NSF Graduate Research Fellowship.}%
\thanks{$^{1}$Department of Mechanical Engineering, Stanford University, Stanford, CA 94305, USA, {\tt\small \{halsted, shorinwa\}@stanford.edu}}%
\thanks{$^{2}$Department of Aeronautics and Astronautics, Stanford University, Stanford, CA 94305, USA
        {\tt\small \{javieryu, schwager\}@stanford.edu}}%
        }
    
\maketitle

\begin{abstract} Distributed optimization provides a framework for deriving distributed algorithms for a variety of multi-robot problems. This tutorial constitutes the first part of a two-part series on distributed optimization applied to multi-robot problems, which seeks to advance the application of distributed optimization in robotics. In this tutorial, we demonstrate that many canonical multi-robot problems can be cast within the distributed optimization framework, such as multi-robot simultaneous localization and { mapping} (SLAM), multi-robot target tracking, and multi-robot task assignment problems. We identify three broad categories of distributed optimization algorithms: distributed first-order methods, distributed sequential convex programming, and the alternating direction method of multipliers (ADMM). We describe the basic algorithmic structure of each category and provide representative algorithms within each category. We then work through a simulation case study of multiple drones collaboratively tracking a ground vehicle.  We compare solutions to this problem using a number of different distributed optimization algorithms. In addition, we implement a distributed optimization algorithm in hardware on a network of Raspberry Pis communicating with XBee modules to illustrate robustness to the challenges of real-world communication networks.

\end{abstract}


\section{Introduction}   
Distributed optimization is the problem of minimizing a joint objective function subject to constraints using an algorithm implemented on a network of communicating computation nodes. In this tutorial, we specifically consider the computation nodes as robots and the network as a multi-robot mesh network.  While distributed optimization has been a longstanding topic of research in the optimization community (e.g., \cite{rockafellar1976monotone, tsitsiklis1984problems}), its usage in multi-robot systems is limited to only a handful of examples. { However, we contend that many problems in multi-robot coordination and collaboration can be formulated and solved within the framework of distributed optimization, yielding a powerful new tool for multi-robot systems.  We show in this tutorial that cooperative estimation \cite{ola2020targettracking}, distributed SLAM, multi-agent learning \cite{wai2018multi}, and collaborative motion planning \cite{bento2013message} are all amenable to approaches based on distributed optimization.}

This tutorial constitutes the first part of a two-part series on distributed optimization methods for multi-robot systems. {In the first part (the tutorial), we focus on introducing the concepts of distributed optimization in application to a broad class of multi-robot problems, the second part (the survey) provides a survey of existing distributed optimization methods, and highlights open research problems in distributed optimization for multi-robot systems. This series is directed towards robotics researchers and practitioners interested in learning about distributed optimization techniques, and their potential to yield novel solutions to problems in multi-robot coordination.}

{We consider problems that are \emph{separable}, meaning the joint objective function can be expressed as a sum over each robot's local objective functions, and the joint constraints can be expressed as the intersection over the robots' local constraints. Each robot only requires knowledge of its own local objective and constraints, and only communicates with one-hop neighbors in a mesh network. The algorithms we discuss are homogeneous, in that each robot executes the same algorithmic steps. There is no specialized leader robot, no hierarchy or differentiated role assignments, and no robot has knowledge of the joint objective or constraints.  In general, these algorithms are iterative, with each robot sharing its intermediate decision variables and/or problem gradients with its one-hop neighbors at each iteration. As the iterations proceed, the decision variables of all the robots converge to a common solution of the optimization problem. In convex problems, each robot obtains a globally optimal solution to the joint problem.  In non-convex problems, the robots typically reach consensus on a locally optimal solution.\footnote{This is the behavior we often observe in practice, although analytical convergence and consensus guarantees for the non-convex case remain an open area of research.}}

{ We describe three broad classes of optimization algorithms: Distributed First-Order Methods (in which the update procedure for the iterates require each robot to compute a gradient of its local objective function), Distributed Sequential Convex Methods (in which the update procedure for the iterates require robots to compute higher-order derivatives, such as Hessians, in addition to gradients), and Alternating Direction Method of Multiplier (ADMM) Methods (in which each robot optimizes a full sub-problem at each iteration).  We give key examples from each class, and discuss their implementation details.  We also implement these algorithms in an example scenario in which multiple aerial robots collaborate to estimate the trajectory of a moving target.  Finally, we demonstrate a hardware implementation of an ADMM algorithm on a network of Raspberry Pis communicating with XBee radios.}

{In some cases, it may not be obvious that a multi-robot problem is of the appropriate form for a distributed optimization algorithm.  One may have to manipulate the problem formulation to express it as a separable optimization.} We demonstrate in this tutorial that many core multi-robot problems, namely multi-robot simultaneous localization and mapping (SLAM), multi-robot target tracking, multi-robot task assignment, collaborative trajectory planning, and multi-robot learning, can be cast in this form. Optimization-based approaches often provide new flexibility, new insights, and new performance guarantees in solving multi-robot problems. For example, multi-robot target tracking problems are typically solved via filtering or smoothing approaches, leading to challenges in managing the cross-correlation of local measurements \cite{ong2008decentralised}. Formulating multi-robot target tracking problems as optimization problems avoids these drawbacks.

{ \subsection{Centralized vs Distributed Optimization} In principle, multi-robot problems can be solved through centralized optimization.  This could be done by passing all information to a leader robot or a base station to perform the computation centrally.  However, such centralized techniques are not scalable to large groups of robots, require large amounts of communication to aggregate the data at one location, and introduce a single point of vulnerability (the leader or base station) to faults and attacks.  Instead, distributed optimization algorithms enable each robot to obtain an optimal solution of the joint problem locally, through communications with one-hop neighbors, without a leader or single point of failure.  

Distributed optimization algorithms also have an inherent data-privacy property.  The robots co-optimize a joint objective without sharing their local ``problem data'' with one another.  Specifically, while robots communicate the value of their local decision variables and/or gradients, they do not expose the functional form of their objective and constraint functions, or directly communication raw sensor data with one another.  This may facilitate cooperation across competing manufacturers or competing service providers without exposing proprietary data, or without violating data privacy laws.

Despite their many advantages, distributed optimization algorithms do come with some drawbacks compared to centralized methods. Since each robot progressively obtains more information via communication with its neighbors, we observe that distributed optimization algorithms require a greater number of iterations for convergence than their centralized counterparts, and often require a longer computation time to converge compared to centralized methods, particularly in small-scale problems. However, there seems to be little research comparing the empirical or theoretical performance of distributed vs centralized optimization algorithms, which presents an interesting direction for future research.  Some distributed algorithms can also be sensitive to hyper-parameter tuning, can have a strong reliance on synchronous algorithmic updates, and can be intolerant of dynamically changing networks. In this tutorial we highlight which algorithm classes suffer from these challenges, and discuss practical ways to accommodate these requirements in robotics problems.}

    \subsection{Contributions}
    This tutorial paper has three primary objectives:
    \begin{enumerate}
        \item Describe three main classes of distributed optimization algorithms.
        \item Highlight the practical implications of typical assumptions made by distributed optimization algorithms and provide potential strategies for addressing the associated challenges.
        \item Demonstrate the formulation of many canonical multi-robot problems as distributed optimization problems.
        \item Provide a case study comparing multiple different distributed optimization algorithms in multi-drone target tracking scenario, both in simulation and on networking hardware.
    \end{enumerate}
    
    \subsection{Organization}
     We present notation and mathematical preliminaries in Sec.~\ref{sec:preliminaries} and formulate the general separable distributed optimization problem in Sec.~\ref{sec:problem_formulation}. Section \ref{sec:distributed_algorithms} describes the three main categories of distributed optimization algorithms and provides representative algorithms for each category.  In Sec.~\ref{sec:multi_robot_optimization_problems}, we demonstrate that many multi-robot problems can be cast within the framework of distributed optimization. { In Sec.~\ref{sec:practical_notes} we offer implementation tips, practical performance observations, and discuss limitations of these methods.} Section \ref{sec:simulations} gives a demonstration of distributed optimization algorithms applied to a multi-drone vehicle tracking problem in simulation and hardware, and we give concluding remarks in Sec.~\ref{sec:conclusion}.

    \section{Notation and Preliminaries}
    \label{sec:preliminaries}
    In this section, we introduce the notation used in this paper and provide the definitions of mathematical concepts relevant to the discussion of the distribution optimization algorithms. We denote the gradient of a function ${f: \mathbb{R}^{n} \rightarrow \mathbb{R}}$ as ${\nabla f}$ and its Hessian as ${\nabla^{2} f}$. We denote the vector containing all ones as $\mathbf{1}_{n}$, where $n$ represents the number of elements in the vector.

    We discuss some relevant notions of the connectivity of a graph.
    
    \begin{definition}[Connectivity of an Undirected Graph]
        An undirected graph $\mathcal{G}$ is connected if a path exists between every pair of vertices $(i,j)$ where ${i,j \in \mathcal{V}}$. Note that such a path might traverse other vertices in $\mathcal{G}$.
    \end{definition}
    
    \begin{definition}[Connectivity of a Directed Graph]
        A directed graph $\mathcal{G}$ is strongly connected if a directed path exists between every pair of vertices $(i,j)$ where ${i,j \in \mathcal{V}}$. In addition, a directed graph $\mathcal{G}$ is weakly connected if the underlying undirected graph is connected. The underlying undirected graph $\mathcal{G}_{u}$ of a directed graph $\mathcal{G}$ refers to a graph with the same set of vertices as $\mathcal{G}$ and a set of edges obtained by considering each edge in $\mathcal{G}$ as a bi-directional edge. Consequently, every strongly connected directed graph is weakly connected; however, the converse is not true.
    \end{definition}
    
    \begin{definition}[Stochastic Matrix]
        A non-negative matrix ${W \in \mathbb{R}^{n \times n}}$ is referred to as a row-stochastic matrix if
        \begin{equation}
            W \mathbf{1}_{n} = \mathbf{1}_{n},
        \end{equation}
        in other words, the sum of all elements in each row of the matrix equals one. 
        We refer to $W$ as a column-stochastic matrix if
        \begin{equation}
            \mathbf{1}_{n}^{\top}W = \mathbf{1}_{n}^\top.
        \end{equation}
        Likewise, for a doubly-stochastic matrix $W$,
        \begin{equation}
            W \mathbf{1}_{n} = \mathbf{1}_{n}\ \text{and}\ \mathbf{1}_{n}^{\top}W = \mathbf{1}_{n}^\top.
        \end{equation}
    \end{definition}

	In distributed optimization in multi-robot systems, robots perform communication and computation steps to minimize some joint objective function. We focus on problems in which the robots' exchange of information must respect the topology of an underlying distributed communication graph, which could possibly change over time.  This communication graph, denoted as $\mc{G}(t) = (\mc{V}(t), \mc{E}(t))$, consists of vertices $\mc{V}(t) = \{1, \dots, N\}$ and edges $\mc{E}(t) \subseteq \mc{V}(t) \times \mc{V}(t)$ over which pairwise communication can occur. For undirected graphs, we denote the set of neighbors of robot $i$ as $\mathcal{N}_{i}(t)$. For directed graphs, we refer to the set of robots which can \textit{send} information to robot $i$ as the set of in-neighbors of robot $i$, denoted by $\mathcal{N}_{i}^{+}(t)$. Likewise, for directed graphs, we refer to the set of robots which can \textit{receive} information from robot $i$ as the out-neighbors of robot $i$, denoted by $\mathcal{N}_{i}^{-}(t)$.

	\section{Problem Formulation}
	\label{sec:problem_formulation}

	{ We consider a general \emph{separable} distributed optimization problem of the form 
 \begin{align}
    	\label{eq:general_problem}
    	\begin{split}
    	\min_x \:&\sum_{i \in \mc{V}} f_i(x)\\
    	\text{subject to }&g_i(x) = 0 \quad \forall i \in \mc{V}
    	\\& h_i(x) \le 0 \quad \forall i \in \mc{V}
    	\end{split}
\end{align}	
 where ${x \in \mathbb{R}^{n}}$ denotes the joint optimization variable, $f_i : \mathbb{R}^{n} \rightarrow \mathbb{R}$ is the local objective function for robot $i$, $g_{i}: \mathbb{R}^{n} \rightarrow \mathbb{R}$ is the equality constraint function of robot $i$, and $h_{i}: \mathbb{R}^{n} \rightarrow \mathbb{R}$ denotes its inequality constraint function. Each robot $i \in \mc{V}$ has access to its local objective constraint functions, but has no knowledge of the local objective and constraint functions of other robots. Such problems arise in many robotics applications where the local objective functions depend on data collected locally by each robot, often in the form of measurements taken by sensors attached to the robot.} The robots seek to collectively solve this joint optimization problem without a leader or central coordinator. We note that not all robots need to have a local constraint function. In these cases, the corresponding constraint functions are omitted in \eqref{eq:general_problem}. 
	
We consider distributed algorithms in which each robot maintains a local copy of the optimization variable, with $x_{i}$ denoting robot $i$'s local vector of optimization variables. Distributed optimization algorithms solve an equivalent reformulation of the optimization problem \eqref{eq:general_problem}, given by
    \begin{align}
    	\label{eq:general_problem_with_agreement_explicit}
    	\begin{split}
    	\min_{\{x_i,\ \forall i \in \mc{V}\}} \:&\sum_{i \in \mc{V}} f_i(x_i)\\
    	\text{subject to }& x_i = x_j \quad \forall (i, j) \in \mc{E} \\&g_i(x_i) = 0 \quad \forall i \in \mc{V}
    	\\& h_i(x_i) \le 0 \quad \forall i \in \mc{V}.
    	\end{split}
	\end{align}
	We call the $x_i = x_j \quad \forall (i,j)\in \mathcal{E}$ the consensus constraints.  %
    Under the assumption that the communication graph is connected for undirected graphs and weakly connected for directed graphs, the optimal cost in \eqref{eq:general_problem_with_agreement_explicit} is equivalent to that in \eqref{eq:general_problem}, and the minimizing arguments $x_i^*$ in \eqref{eq:general_problem_with_agreement_explicit} are equal to the minimizing argument $x^*$ of \eqref{eq:general_problem} for all robots $i = 1, \ldots,n$. To simplify notation, we introduce the set ${\mathcal{X}_{i} = \{x_{i} \mid g_i(x_i) = 0, h_i(x_i) \le 0\}}$, representing the feasible set given the constraint functions $g_{i}$ and $h_{i}$. Consequently, we can express the problem in \eqref{eq:general_problem_with_agreement_explicit} succinctly as follows:
    \begin{align}
    	\label{eq:general_problem_with_agreement}
    	\begin{split}
    	\min_{\{x_i \in \mathcal{X}_{i},\ \forall i \in \mc{V}\}} \:&\sum_{i \in \mc{V}} f_i(x_i)\\
    	\text{subject to }& x_i = x_j \quad \forall (i, j) \in \mc{E}.
    	\end{split}
    \end{align}

	\section{Classes of Distributed Optimization Algorithms}
	\label{sec:distributed_algorithms}
	In this section, we categorize distributed optimization algorithms into three broad classes --- Distributed First-Order Methods, Distributed Sequential Convex Programming, and ADMM Methods --- { based on shared mechanisms for achieving convergence (and not necessarily based on their applicability to multi-robot problems)}.  We provide a brief overview of each category, by considering a representative distributed algorithm within each category. In the subsequent discussion, we consider the separable optimization problem in \eqref{eq:general_problem_with_agreement}.
	
    Before describing the specific algorithms that solve distributed optimization problems, we first consider the general framework that all of these approaches share.  Each algorithm progresses over discrete iterations $k = 0, 1, \dots$ until convergence. In general, each iteration consists of a communication step and a computation step. Besides assuming that each robot has the sole capability of evaluating its local objective function $f_i$, we also distinguish between the ``internal'' variables $\mc{P}_i^{(k)}$ that the robot computes at each iteration $k$ and the ``communicated'' variables $\mc{Q}_i^{(k)}$ that the robot communicates to its neighbors. Each algorithm also involves parameters $\mc{R}_i^{(k)}$, which generally require coordination among all of the robots, but can typically be assigned before deployment of the system.
    
    In distributed optimization, all the robots seek to collectively minimize the joint objective function in \eqref{eq:general_problem_with_agreement} while achieving consensus on a common set of minimizing optimization variables. { Each of the three class we describe treats the consensus constraints in \eqref{eq:general_problem_with_agreement} differently.  In distributed first order methods, from the perspective of a single robot, the update iterations
represent a trade-off between optimality of a robot's individual solution based on its local
objective function versus reaching agreement with its neighbors, either on the decision variable directly, or on the gradient of the global objective.  Asymptotically the robots' decision variables or gradient converge to a consensus leading to global optimality for convex problems. In distributed sequential convex methods, individual robots use communication to build approximate global Hessians and gradients to execute approximate second order update steps, asymptotically leading each agent to obtain a global minimum in the convex case.  Finally, for the alternating direction method of multipliers these consensus constraints are enforced explicitly through an augmented Lagrangian constrained optimization approach. The key insight underlying this approach is that minimizing the local objective functions
subject to these additional agreement constraints is equivalent to minimizing the joint objective function over a collective decision variable.}
	
	\subsection{Distributed First-Order Methods}
	{ Gradient decent methods have been widely applied to solve broad classes of optimization problems, particularly unconstrained problems. To simplify the discussion of these methods, we consider the unconstrained variant of \eqref{eq:general_problem_with_agreement}, where we only retain the consensus constraints, and disregard the constraint functions $g_i(x_i)$ and $h_i(x_i)$. We note that extensions of gradient descent to constrained optimization typically involve a projection of the iterates to the feasible set, a method known as \emph{projected gradient descent}. In the second part of our series \cite{shorinwa_distributed_2023-1}, we discuss extensions of gradient descent methods to constrained optimization in greater detail.} In general, gradient descent methods only require the computation of the gradient (i.e., the first derivative of the objective and constraint functions); hence, these methods are also referred to as \emph{first-order} methods. When applied to the unconstrained joint optimization problem, the updates to the optimization variable take the form
	\begin{equation}
	    \label{eqn:cent_grad_desc}
	    x^{(k+1)} = x^{(k)} - \alpha^{(k)} \nabla f(x^{(k)})
	\end{equation}
	where $\alpha^{(k)}$ denotes a diminishing step-size and $\nabla f(x^{(k)})$ denotes the gradient of the objective function, given by
	\begin{equation}
	    \label{eq:centralized_gradient}
	    \nabla f(x) = \sum_{i \in \mathcal{V}} \nabla f_i(x).
	\end{equation}
	From \eqref{eq:centralized_gradient}, computation of $\nabla f(x)$ requires knowledge of the objective function of all robots, which is unavailable to any individual robot, and thus requires aggregation of this information at a central node.
	
	Distributed First-Order (DFO) algorithms circumvent this underlying challenge by enabling each robot to utilize only its local gradients, while communicating with its neighbors to reach consensus on a common solution. { In many DFO methods, a robot aggregates the information of its neighbors by taking the weighted combination of the local variables or gradients as specified by a stochastic weighting matrix $W$. The stochastic matrix $W$ must be compatible with the underlying communication network (i.e., $w_{ij}$ is only non-zero if robot $j$
can send information to robot $i$).

We begin with a basic distributed gradient descent method, described by the update procedure:
    \begin{equation}
        x_i^{(k+1)} = \sum_{j \in \mc{N}_i \cup \{i\}} w_{ij}x_j^{(k)} - \alpha^{(k)} \nabla f_i\left(x_i^{(k)}\right), \label{eqn:CTA}
    \end{equation}
where each robot \emph{mixes} its local estimates with those of its neighbors by taking a weighted combination of these local estimates before taking a step in the direction of its local gradient. More generally, a subgradient ${\partial f_{i}(x_{i}^{(k)})}$ (where $\partial f_{i}$ denotes the subgradient of $f_{i}$) can be utilized in place of the gradient of the local objective function, yielding the canonical distributed subgradient method \cite{nedic2007rate}. This paradigm, consisting of taking a weighted combination of local estimates prior to a descent step, is referred to as the Combine-Then-Adapt (CTA) Paradigm. In contrast, in Adapt-Then-Combine (ATC) methods, each robot updates its local optimization variable using its gradient prior to combining its local variable with that of its neighbors, with the update procedure given by
	\begin{equation}
	    x_i^{(k+1)} = \sum_{j \in \mc{N}_i \cup \{i\}} w_{ij}\left(x_j^{(k)} - \alpha^{(k)}\nabla f_j\left(x_j^{(k)}\right)\right), \label{eqn:ATC}
	\end{equation}
	{ where ${x_{j}^{(k)} \in \mathbb{R}^{n}}$ denotes the local variable of neighboring robot $j$, and each robot updates its local variable $x_i^{(k+1)}$ using the local gradient before communicating its local variable with its neighbors and aggregating their respective updates.}
    Consequently, we can further categorize DFO methods into two broad subclasses: Adapt-Then-Combine (ATC) methods and Combine-Then-Adapt (CTA) methods, based on the relative order of the communication and computation procedures.

    The algorithms given by \eqref{eqn:ATC} and \eqref{eqn:CTA} do not converge to the optimal solution of the joint optimization problem, in general. To see this, consider the case where ${x_{i} = x^{\star}}$,~${\forall i \in \mathcal{V}}$, where ${x^{\star}}$ denotes the optimal solution of the joint optimization problem. In the ATC approach, we can express the update procedure as the difference between two terms: ${\sum_{j \in \mc{N}_i \cup \{i\}} w_{ij} x_j^{(k)}}$ and ${\sum_{j \in \mc{N}_i \cup \{i\}} w_{ij} \alpha^{(k)}\nabla f_j\left(x_j^{(k)}\right)}$. Given that $W$ is row-stochastic, the first term in ATC and CTA approaches simplifies to ${x^{\star}}$. However, in ATC approaches, the second term represents a weighted combination of the local gradients of each robot, which is not necessarily zero. In fact, we only have ${\sum_{i \in \mathcal{V}} \nabla f_{i}(x^{\star}) = 0}$, in the general case. Likewise, in CTA methods, the second term represents the local gradient of each agent, which is not necessarily zero. As a result, the iterate $x_i^{(k+1)}$ moves away from the optimal solution $x^{\star}$.

     If $\alpha^{(k)}$ did not asymptotically converge to zero, then the iterates would only converge to a neighborhood of the globally optimal value (observe that substituting the optimal value into \eqref{eqn:ATC} or \eqref{eqn:CTA} yields a nonzero innovation) \cite{nedic2009}. {
     If the step-size satisfies the conditions: ${\sum_{k = 0}^{\infty} \alpha(k) = \infty}$ and ${\sum_{k = 0}^{\infty} (\alpha(k))^{2} < \infty}$, then convergence of the iterates to an optimal solution is guaranteed \cite{nedic2009distributed, lobel2010distributed}. An example of a step-size rule satisfying these conditions is given by ${\alpha^{(k)} = \frac{\alpha^{(0)}}{k}}$. Although both conditions are \emph{sufficient} for convergence, only the non-summable condition is \emph{necessary} \cite{chen2012fast}. In practice, an optimal diminishing step-size is given by ${\alpha^{(k)} = \frac{\alpha^{(0)}}{\sqrt{k}}}$, which is not square-summable \cite{nedic2014distributed, chen2012fast}.}
	
    { In extensions of these basic approaches, we replace the gradient $\nabla f_i\left(x_i^{(k)}\right)$ with a new variable $y_i^{(k)}$ that uses consensus to aggregate gradient information from the other robots and track the average gradient of the joint objective function.} Gradient tracking methods, for example DIGing \cite{nedic2017digging}, employ an estimate of the average gradient computed through dynamic average consensus with 
	\begin{equation}
	    {y_i^{(k+1)} = \sum_{j \in \mc{N}_i \cup \{i\}} w_{ij} y_j^{(k)} + \left[\nabla f_i\left(x_i^{(k+1)}\right) - \nabla f_i\left(x_i^{(k)}\right)\right]}. \label{eqn:DIGing}
	\end{equation}
 { The iterate $x_i^{(k)}$ of each agent is guaranteed to converge to the optimal solution $x^{\star}$ under a constant step-size, provided the communication network is connected and certain other conditions on the network topology and the objective functions hold \cite{nedic2017digging}. Moreover, the iterate $y_i^{(k)}$ converges to the average gradient of the individual objective functions \cite{zhu2010discrete}, given convergence of $x_i^{(k)}$ to the limit point $x^{\star}$.
 }
 At initialization of the algorithm, all the robots select a common step-size. Further, robot $i$ initializes its local variables with ${x_{i}^{(0)} \in \mathbb{R}^{n}}$ and ${y_{i}^{(0)} = \partial f_{i}(x_{i}^{(0)})}$. Algorithm \ref{alg:DIGing} summarizes the update procedures in the distributed gradient tracking method DIGing \cite{nedic2017digging}. Other gradient tracking algorithms include \cite{shi2015extra, li2019decentralized, xi2017add}. We note that ATC methods, e.g., \cite{xu2015augmented}, are compatible with uncoordinated step-sizes, i.e., each robot does not have to use the same step-size. Unlike ATC methods, CTA methods require a common step-size among the robots for convergence to an optimal solution. Moreover, distributed gradient tracking has been extended to the conjugate gradient setting, where the update procedures are defined using conjugate gradients rather than the gradient of the objective function, for faster convergence \cite{shorinwa2024conjugate}.

    \begin{algorithm}[t]
    	\caption{DIGing}\label{alg:DIGing}
    	\textbf{Initialization:} $k \gets 0$, ${x_{i}^{(0)} \in \mathbb{R}^{n}}$, ${y_{i}^{(0)} = \nabla f_{i}(x_{i}^{(0)})}$
    	\;
        \textbf{Internal variables:} $\mc{P}_i^{(k)} = \emptyset$
        \;
        \textbf{Communicated variables:} $\mc{Q}_i^{(k)} = \left(x_i^{(k)}, y_i^{k}\right)$
        \;
        \textbf{Parameters:} $\mc{R}_i^{(k)} = \left(\alpha, w_i\right)$\;
        
        \doparallel ($\forall i \in \mc{V}$){stopping criterion is not satisfied} {
    	    Communicate $\mc{Q}_i^{(k)}$ to all $j \in \mc{N}_i$\;
    	    Receive $\mc{Q}_j^{(k)}$ from all $j \in \mc{N}_i$
    	    \begin{align*}
        	    x_i^{(k+1)} &= \sum_{j \in \mc{N}_i \cup \{i\}} w_{ij}x_j^{(k)} - \alpha y_i^{(k)} \\
        	    y_i^{(k+1)} &= \sum_{j \in \mc{N}_i \cup \{i\}} w_{ij} y_j^{(k)} + \nabla f_i(x_i^{(k+1)}) - \nabla f_i(x_i^{(k)})
    	    \end{align*}
    	    
    	    $k \gets k + 1$
        }
    \end{algorithm}

	\subsection{Distributed Sequential Convex Programming}
    Sequential convex programming entails solving an optimization problem by computing a sequence of iterates, representing the solution of a series of approximations of the original problem. Newton's method is a prime example of a sequential convex programming method. In Newton's method, and more generally, quasi-Newton methods, we take a quadratic approximation of the objective function at an operating point $x^{(k)}$, resulting in
    \begin{equation}
        \begin{aligned}
            \tilde{f}(x) &= f(x^{(k)}) + \nabla f(x^{(k)})^{\top} (x - x^{(k)}) \\
            & \quad + \frac{1}{2} (x - x^{(k)})^{\top}H(x^{(k)})(x - x^{(k)}),
        \end{aligned}
    \end{equation}
    where $H(\cdot)$ denotes the Hessian of the objective function, $\nabla^{2} f$, or its approximation. Subsequently, we compute a solution to the quadratic program, given by
    \begin{equation}
        x^{(k + 1)} = x^{(k)} - H\bigl(x^{(k)}\bigr)^{-1} \nabla \tilde{f}(x^{(k)}),
    \end{equation}
    which requires centralized evaluation of the gradient and Hessian of the objective function. Distributed Sequential Programming enable each robot to compute a local estimate of the gradient and Hessian of the objective function, and thus allows for the local execution of the update procedures. We consider the NEXT algorithm \cite{di2016next} to illustrate this class of distributed optimization algorithms. We assume that each robot uses a quadratic approximation of the optimization problem as its convex surrogate model $U(\cdot)$. In NEXT, each robot maintains an estimate of the average gradient of the objective function, as well as an estimate of the gradient of the objective function excluding its local component (i.e., the gradient of ${\sum_{j \neq i} f_{j}(x_{i})}$ for robot $i$, { which we denote by $\ti{\pi}_i^{(k)}$}). At a current iterate ${x_i^{(k)}}$, robot $i$ creates a quadratic approximation of the optimization problem, given by
    \begin{equation}
        \label{eq:convex_approximation}
            \begin{aligned}
                \underset{\tilde{x}_{i} \in \mathcal{X}_{i}}{\mathrm{minimize}} &\left(\nabla f_{i}(x_{i}^{(k)}) + \ti{\pi}_i^{(k)}\right)^{\top} \bigl(\tilde{x}_{i} - x_{i}^{(k)}\bigr) \\
                & \quad + \frac{1}{2} \bigl(\tilde{x}_{i} - x_{i}^{(k)}\bigr)^{\top}H_{i}\bigl(x_{i}^{(k)}\bigr)\bigl(\tilde{x}_{i} - x_{i}^{(k)}\bigr),
            \end{aligned}
    \end{equation}
   which { takes into account the robot's local Hessian $H_i$ or its estimate (e.g., computed using a quasi-Newton update scheme \cite{dennis1977quasi, byrd1996analysis, tapia1988secant}) and can be solved locally}. Each robot computes a weighted combination of its current iterate and the solution of \eqref{eq:convex_approximation}, given by the procedure
	\begin{equation}
	    z_i^{(k)} = x_i^{(k)} + \alpha^{(k)} \left(\ti{x}_i^{(k)} - x_i^{(k)} \right),
	\end{equation}
	where ${\alpha^{(k)} \in (0, 1)}$ denotes a diminishing step-size. Subsequently, robot $i$ computes its next iterate by taking a weighted combination of its local estimate $z_i^{(k)}$ with that of its neighbors via the procedure
	\begin{equation}
	    x_i^{(k + 1)} = \sum_{j \in \mc{N}_{i} \cup \{i\}} w_{ij} z_j^{(k)},
	\end{equation}
	for consensus on a common solution of the original optimization problem, where the weight $w_{i,j}$ must be compatible with the underlying communication network. In addition, robot $i$ updates its estimates of the average gradient of the objective function, denoted by $y_i$, { using dynamic average consensus in the same form as \eqref{eqn:DIGing}}.
{ Updating $\ti{\pi}_i^{(k)}$ takes a similar form. In the limit that the iterates approach a common value $x^*$, $y_i$ approaches the average gradient of the joint objective function at $x^*$ and so does $\ti{\pi}_i^{(k)} + \nabla f_i\left(x_i^{(k)}\right)$. Thus, NEXT reasons that an appropriate update for $\ti{\pi}_i$ takes the following form:}
	\begin{equation}
	    \ti{\pi}_i^{(k+1)} = N \cdot y_i^{(k + 1)} - \nabla f_i(x_i^{(k+1)}).
	\end{equation}
	Each agent initializes its local variables with ${x_{i}^{(0)} \in \mathbb{R}^{n}}$, ${y_{i}^{(0)} = \nabla f_{i}(x_{i}^{(0)})}$, and ${\ti{\pi}_i^{(k+1)} = N y_{i}^{(0)} - \nabla f_{i}(x_{i}^{(0)})}$, prior to executing the above update procedures. {We note that NEXT is guaranteed to converge to a stationary point of the optimization problem \cite{di2016next}}. Algorithm \ref{alg:NEXT} summarizes the update procedures in NEXT \cite{di2016next}.

 { Other algorithms that use distributed sequential convex programming include methods that perform distributed Newton's method \cite{Mokhtari2015}, and distributed quasi-Newton methods \cite{Eisen2017, shorinwa2024distributedquasi}. Furthermore, algorithms that use consensus on local Hessians exist \cite{liu2023communication}, often at the expense of greater communication overhead.}

    \begin{algorithm}[t]
    	\caption{NEXT}\label{alg:NEXT}
    	\textbf{Initialization:} $k \gets 0$, ${x_{i}^{(0)} \in \mathbb{R}^{n}}$, ${y_{i}^{(0)} = \nabla f_{i}(x_{i}^{(0)})}$, {\phantom{Initialization:}} ${\ti{\pi}_i^{(0)} = N y_{i}^{(0)} - \nabla f_{i}(x_{i}^{(0)})}$
    	\;
        \textbf{Internal variables:} $\mc{P}_i = \left(x_i^{(k)}, \tilde{x}_i^{(k)}, \ti{\pi}_i^{(k)} \right)$
        \;
        \textbf{Communicated variables:} $\mc{Q}_i^{(k)} = \left(z_i^{(k)}, y_i^{(k)} \right)$
        \;
        \textbf{Parameters:} $\mc{R}_i^{(k)} = \left(\alpha^{(k)}, w_i, U(\cdot), \mc{X}_{i} \right)$\;
        
        \doparallel ($\forall i \in \mc{V}$){stopping criterion is not satisfied} {
            \begingroup\abovedisplayskip=0pt
            \begin{flalign*}
                \ti{x}_i^{(k)} &= \argmin_{x \in \mc{X}_{i}}\ U \left( x; x_i^{(k)}, \ti{\pi}_i^{(k)} \right) &\\
                z_i^{(k)} &= x_i^{(k)} + \alpha^{(k)} \left(\ti{x}_i^{(k)} - x_i^{(k)} \right) &
            \end{flalign*}
            \endgroup
    	    Communicate $\mc{Q}_i^{(k)}$ to all $j \in \mc{N}_i$\;
    	    Receive $\mc{Q}_j^{(k)}$ from all $j \in \mc{N}_i$
    	    \begin{flalign*}
                x_i^{(k+1)} &= \sum_{j \in \mc{N}_{i} \cup \{i\}} w_{ij} z_j^{(k)} &\\
                y_i^{(k+1)} &= \sum_{j \in \mc{N}_{i} \cup \{i\}} w_{ij} y_j^{(k)} &\\
                & \qquad + \left[ \nabla f_i(x_i^{(k + 1)}) -  \nabla f_i(x_i^{(k)}) \right] &\\
                \ti{\pi}_i^{(k+1)} &= N \cdot y_i^{(k + 1)} - \nabla f_i(x_i^{(k+1)}) &
            \end{flalign*}
    	    
    	    $k \gets k + 1$
        }
    \end{algorithm}

    \subsection{Alternating Direction Method of Multipliers}
	The alternating direction method of multipliers (ADMM) belongs to the class of optimization algorithms referred to as the method of multipliers (or augmented Lagrangian methods), which compute a primal-dual solution pair of a given optimization problem. The method of multipliers proceeds in an alternating fashion: the primal iterates are updated as minimizers of the augmented Lagrangian, and subsequently, the dual iterates are updated via dual (gradient) ascent on the augmented Lagrangian. The procedure continues iteratively until convergence or termination. The augmented Lagrangian of the problem in \eqref{eq:general_problem_with_agreement} (with only the consensus constraints) is given by
	\begin{equation}
	    \label{eq:augmented_lagrangian}
	    \begin{aligned}
        	\mathcal{L}_{a}(\mathbf{x},q) &=  \sum_{i = 1}^{N} f_{i}(x_{i}) \\
        	& \enspace + \sum_{i = 1}^{N} \sum_{j \in \mathcal{N}_{i}} \left(q_{i,j}^{\top}(x_{i} - x_{j}) +  \frac{\rho}{2}  \norm{x_{i} - x_{j}}_{2}^{2}\right),
    	\end{aligned}
	\end{equation}
	where $q_{i,j}$ represents a dual variable for the consensus constraints between robots $i$ and $j$, ${q = \left[q_{i,j}^{\top},\ \forall (i,j) \in \mathcal{E}\right]^{\top}}$, and ${\mathbf{x} = \left[x_{1}^{\top},x_{2}^{\top},\cdots,x_{N}^{\top}\right]^{\top}}$. The parameter ${\rho > 0}$ represents a penalty term on the violations of the consensus constraints. Generally, the method of multipliers computes the minimizer of the augmented Lagrangian with respect to the joint set of optimization variables, which hinders distributed computation. In contrast, in the alternating direction method of multipliers, the minimization procedure is performed block-component-wise, enabling parallel, distributed computation of the minimization subproblem in the consensus problem. However, many ADMM algorithms still require some centralized computation, rendering them not fully-distributed in multi-robot mesh network sense that we consider in this paper. 
	
	We focus here on ADMM algorithms that are distributed over robots in a mesh network, with each robot executing the same set of distributed steps.  We specifically consider the consensus alternating direction method of multipliers (C-ADMM) \cite{mateos2010distributed} as a representative algorithm within this category. C-ADMM introduces auxiliary optimization variables into the consensus constraints in \eqref{eq:general_problem_with_agreement} to enable fully-distributed update procedures. The primal update procedure of robot $i$ takes the form
	\begin{equation}
	    \label{eq:c_admm_primal_update}
	    \begin{aligned}
    	    x_i^{(k+1)} &= \argmin_{x_i \in \mathcal{X}_{i}} \Bigg\{f_i(x_i) + x_i^\top y_i^{(k)} \\ 
    	    &\hspace{5em} + \rho \sum_{j \in \mc{N}_i} \left\Vert x_i - \frac{1}{2}\left(x_i^{(k)} + x_j^{(k)}\right)\right\Vert^2_{2} \Bigg\},
	    \end{aligned}
	\end{equation}
	which only requires information locally available to robot $i$, including information received from its neighbors (i.e., ${x_{j}^{k},\ \forall j \in \mathcal{N}_{i}}$). As a result, this procedure can be executed locally by each agent, in parallel. After communicating with its neighbors, each robot updates its local dual variable using the procedure
	\begin{equation}
	    y_i^{(k+1)} = y_i^{(k)} + \rho\sum_{j \in \mc{N}_i} \left(x_i^{(k + 1)} - x_j^{(k + 1)}\right),
	\end{equation}
	where $y_i$ denotes the composite dual variable of robot $i$, corresponding to the consensus constraints between robot $i$ and its neighbors, which is initialized to zero. Algorithm \ref{alg:C-ADMM} summarizes the update procedures in C-ADMM \cite{mateos2010distributed}.

    \begin{algorithm}[t]
    	\caption{C-ADMM} \label{alg:C-ADMM}
    	\textbf{Initialization:} $k \gets 0$, ${x_{i}^{(0)} \in \mathbb{R}^{n}}$, ${y_{i}^{(0)} = 0}$
    	\;
        \textbf{Internal variables:} $\mc{P}_i^{(k)} = y_i^{(k)}$
        \;
        \textbf{Communicated variables:} $\mc{Q}_i^{(k)} = x_i^{(k)}$
        \;
        \textbf{Parameters:} $\mc{R}_i^{(k)} = \rho$\;
        
        \doparallel ($\forall i \in \mc{V}$){stopping criterion is not satisfied} {
            \begingroup\abovedisplayskip=0pt
            \begin{flalign*} 
                &x_i^{(k+1)} = \argmin_{x_i \in \mathcal{X}_{i}} \Bigg\{f_i(x_i) + x_i^\top y_i^{(k)} \cdots \\
                & \hspace{6em} + \rho \sum_{j \in \mc{N}_i} \left\Vert x_i - \frac{1}{2}\left(x_i^{(k)} + x_j^{(k)}\right)\right\Vert^2_{2} \Bigg\} &
            \end{flalign*}
            \endgroup
    	    Communicate $\mc{Q}_i^{(k)}$ to all $j \in \mc{N}_i$\;
    	    Receive $\mc{Q}_j^{(k)}$ from all $j \in \mc{N}_i$
    	    \begin{flalign*}
                y_i^{(k+1)} &= y_i^{(k)} + \rho\sum_{j \in \mc{N}_i} \left(x_i^{(k + 1)} - x_j^{(k + 1)}\right) &
            \end{flalign*}
    	    
    	    $k \gets k + 1$
        }
    \end{algorithm}

    {
    \subsection*{Synopsis}
    We summarize the notable features of each category of distributed algorithms in Table~\ref{tab:algorithm_summary}, which should be considered when selecting a distributed algorithm for a multi-robot problem. In general, the update procedures in distributed first-order (DFO) algorithms require lower-complexity computational operations, which makes them suitable for problems where each robot has limited access to computational resources \cite{nedic2009, shi2015extra, nedic2017digging}. Further, DFO algorithms accommodate dynamic, unidirectional and bidirectional communication networks. However, DFO algorithms are generally not amenable to constrained problems, limiting their applications in some multi-robot problems. On the other hand, while distributed sequential programming (DSQP) algorithms are suitable for problems with dynamic bidirectional communication networks, these algorithms do not generally extend to unidirectional networks \cite{Mokhtari2015, Eisen2017}. In addition, while some DSQP algorithms \cite{di2016next, tian2018asy} are suitable for constrained optimization, this is not the case for all methods of this class. In contrast, although distributed algorithms based on the alternating direction method of multipliers (ADMM) do not address dynamic, unidirectional communication networks, ADMM-based algorithms apply to constrained optimization \cite{mateos2010distributed, ola2020SOVA}. Moreover, ADMM-based algorithms show better robustness to the selection of algorithm parameters such as the step-size or penalty parameter. However, ADMM-based methods incur a greater computational overhead, as the optimization subproblems arising in the update procedures do not necessarily have closed-form solutions.
    }

    \begin{table*}[th]
    	\centering
    	\caption{Suitable distributed optimization algorithms for different complicating attributes common in multi-robot problems. The information displayed is based on the representative algorithm (indicated by the citation) considered in each algorithm class.}
    	\label{tab:algorithm_summary}
    	\begin{adjustbox}{width=0.8\linewidth}
    		{\begin{tabular}{c c c c}
    			\toprule
    			Attribute & DFO (e.g., \cite{nedic2017digging}) & DSCP (e.g., \cite{di2016next}) & ADMM (e.g., \cite{mateos2010distributed}) \\
    			\midrule
    			Dynamic Communication Networks & \cmark & \cmark & \xmark \\
    			Lossy Communication & \cmark & \cmark & \xmark \\
    			Unidirectional Communication Networks & \cmark & \xmark & \xmark \\
    			Bidirectional Communication Networks & \cmark & \cmark & \cmark \\
                    Constrained Problems & \xmark & \cmark & \cmark \\
                    Robustness to Step-Size/Penalty-Parameter & \xmark & \xmark & \cmark \\
    			\bottomrule
    		\end{tabular}}
    	\end{adjustbox}
    \end{table*}

 	\section{Multi-Robot Problems Posed as Distributed Optimizations}
	\label{sec:multi_robot_optimization_problems}
    { Many robotics problems have a distributed structure, although this structure might not be immediately apparent. In many cases, applying distributed optimization methods requires reformulating the original problem into a separable form that allows for distributed computation of the problem variables locally by each robot.}
    In this section, we consider five general problem categories that can be solved using distributed optimization tools: multi-robot SLAM, multi-robot target tracking, multi-robot task assignment, collaborative planning, and multi-robot learning. We note that an optimization-based approach to solving some of these problems might not be immediately obvious. However, we show that many of these problems can be quite easily formulated as distributed optimization problems through the introduction of auxiliary optimization variables, in addition to an appropriate set of consensus constraints.

    \subsection{Multi-Robot Simultaneous Localization and Mapping (SLAM)}
    In multi-robot simultaneous localization and mapping (SLAM) problems, a group of robots seek to estimate their position and orientation (pose) within a consistent representation of their environment.  In a full landmark-based SLAM approach, {{we consider optimizing over both $M$ map features $m_1, \dots, m_M$ as well as $N$ robot poses $x_1, \dots, x_N$ over a duration of ${T + 1}$ timesteps:}}
     \begin{equation}
        \label{eq:pose_graph_slam_problem}
        \begin{aligned}
            \underset{\mathbf{x},m}{\mathrm{minimize}} &\sum_{i=1}^{N} \sum_{t=0}^{T - 1} \norm{\bar{z}_{i,t}(x_{i,t},x_{i,t+1}) - \hat{z}_{i,t + 1}}_{\Omega_{i,t}}^{2} \\
            &+ \sum_{i = 1}^{N} \sum_{k=1}^{M} \norm{\tilde{z}_{i}^{k}(x_{i}, m_{k}) - \Breve{z}_{i}^{k}}_{\Lambda_{i,t}}^{2}.
        \end{aligned}
    \end{equation}
     { The $z$ terms denote measurements ($\hat{z}$, $\Breve{z}$) and measurement functions ($\bar{z}$, $\tilde{z}$):} the expected relative poses $\bar{z}_{i,t}$ are functions of two adjacent poses of robot $i$ derived from robot odometry measurements, and the expected relative pose $\tilde{z}_{i}^{k}$ is a function of the pose of robot $i$ and the position of map feature $k$. We have concatenated the problem variables in \eqref{eq:pose_graph_slam_problem}, with ${x_{i} = \left[x_{i,0}^{\top},x_{i,1}^{\top},\cdots,x_{i,T}^{\top}\right]^{\top}}$, ${\mathbf{x} = \left[x_{1}^{\top},x_{2}^{\top},\cdots,x_{N}^{\top}\right]^{\top}}$, and ${m = \left[m_{1}^{\top},m_{2}^{\top},\cdots,m_{M}^{\top}\right]^{\top}}$. The error terms in the objective function are weighted by the information matrices $\Omega_{i,t}$ and $\Lambda_{i,t}$ associated with the measurements collected by robot $i$.

	\begin{figure}[tb]
        \centering
        \includegraphics[width=0.8\linewidth]{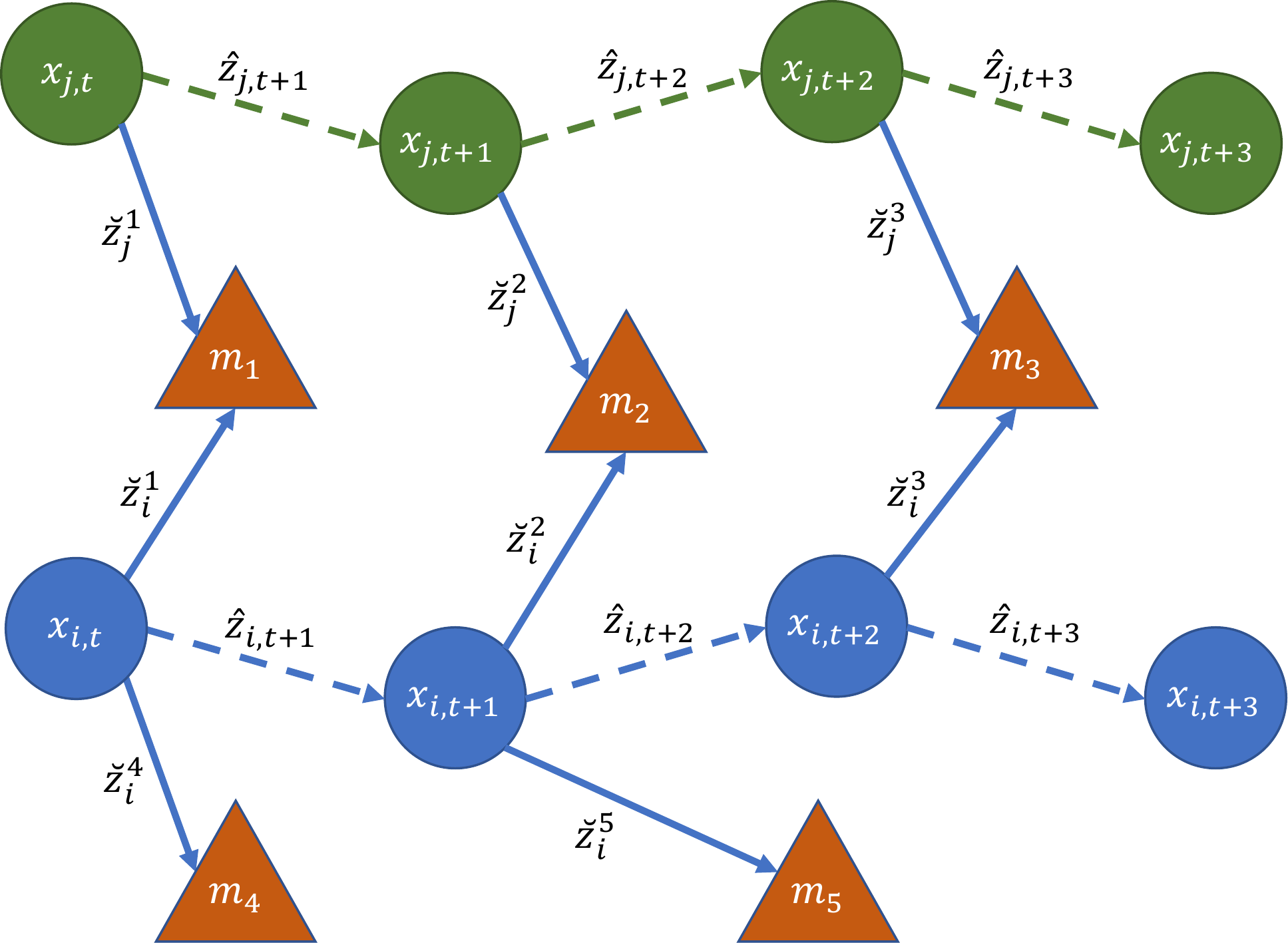}
        \caption{A factor graph representation of a multi-robot SLAM problem, where two robots, robot $i$ (blue circles) and $j$ (green circles), seek to jointly estimate a set of map features ${\{m_{1},m_{2},\cdots\}}$ (orange triangles) in addition to their own pose trajectory ${\{x_{i,t},x_{j,t},\ \forall t\}}$, from the set of odometry measurements ${\{\hat{z}_{i,t},\hat{z}_{j,t}\}}$ and observations of each map feature $k$ ${\{\Breve{z}_{i}^{k}, \Breve{z}_{j}^{k}\}}$.}
        \label{fig:pose_graph}
	\end{figure}

    Although the first set of terms in the objective function of the optimization problem \eqref{eq:pose_graph_slam_problem} is separable among the robots, the second set of terms is not. Consequently, the optimization problem must be reformulated. Non-separability of the objective function arises from the coupling between the map features and the robot poses. To achieve separability of the objective function, we can introduce local copies of the variables corresponding to each feature, with an associated set of consensus (equality) constraints to ensure that the resulting problem remains equivalent to the original problem \eqref{eq:pose_graph_slam_problem}. The resulting problem takes the form
    \begin{equation}
        \label{eq:pose_graph_slam_problem_reformulated}
        \begin{aligned}
            \underset{\mathbf{x},\hat{m}_{1},\hat{m}_{2},\cdots,\hat{m}_{N}}{\mathrm{minimize}} &\sum_{i=1}^{N} \sum_{t=0}^{T - 1} \norm{\bar{z}_{i,t}(x_{i,t},x_{i,t+1}) - \hat{z}_{i,t + 1}}_{\Omega_{i,t}}^{2} \\
            &+ \sum_{i = 1}^{N} \sum_{k=1}^{M} \norm{\tilde{z}_{i}^{k}(x_{i}, \hat{m}_{i,k}) - \Breve{z}_{i}^{k}}_{\Lambda_{i,t}}^{2} \\
            \mathrm{subject\ to}\ &\hat{m}_{i} = \hat{m}_{j} \quad \forall (i,j) \in \mathcal{E},
        \end{aligned}
    \end{equation}
    where robot $i$ maintains $\hat{m}_{i}$, its local copy of the map $m$. { We note that $x_i$ is the trajectory of robot $i$ and is only estimated by robot $i$.} The problem \eqref{eq:pose_graph_slam_problem_reformulated} is separable among the robots, { who enforce consensus between their representations of the map}; in other words, its objective function can be expressed in the form
    \begin{equation}
        \begin{aligned}
            f(\mathbf{x},\hat{m}_{1},\hat{m}_{2},\cdots,\hat{m}_{N}) = \sum_{i = 1}^{N} f_{i}(x_{i},\hat{m}_{i}),
        \end{aligned}
    \end{equation}
    where
    \begin{equation}
        \begin{aligned}
            f_{i}(x_{i},\hat{m}_{i}) &= \sum_{t=0}^{T - 1} \norm{\bar{z}_{i,t}(x_{i,t},x_{i,t+1}) - \hat{z}_{i,t + 1}}_{\Omega_{i,t}}^{2} \\
            &+ \sum_{k=1}^{M} \norm{\tilde{z}_{i}^{k}(x_{i}, \hat{m}_{i,k}) - \Breve{z}_{i}^{k}}_{\Lambda_{i,t}}^{2}
        \end{aligned}.
    \end{equation}
    { Note that the consensus constraints only involve a subset of the local variables of each robot. Distributed optimization algorithms are amenable to problems of this form, without any significant modifications. In methods requiring a weighting matrix, considering robot $i$, only variables involved in the consensus constraints are combined (mixed) with those of its neighbors. Likewise, variants of ADMM, such as SOVA \cite{ola2020SOVA}, can be applied to this problem.}  We can interpret the bundle adjustment problem similarly---in this case, the map features represent the scene geometry and the robot poses include the optical characteristics of the respective cameras.
    However, a challenge in applying this approach in unstructured environments is ensuring that multiple robots agree on the labels of the map landmarks.

    {{
    An alternative approach is pose graph optimization, which avoids explicitly estimating the map by representing the robots' trajectories as a graph in which the edges represent the estimated transformation between poses. %
    A pose $i$ consists of a position (which we represent by the vector $\tau_i$) and orientation (which we represent by the rotation matrix $R_i$).
    In this perspective, the task of determining robot trajectories consists of two stages, performed sequentially.  In the ``front-end,'' the robots process raw sensor measurements to estimate relative poses consisting of a relative rotation ($\tilde{R}_{ij} \approx R_i^{-1}R_j$) and relative translation  ($\tilde{\tau}_{ij} \approx \tau_j - \tau_i$).  The second stage is the ``back-end,'' in which robots find optimal robot poses given those relative pose measurements.
    Under the assumption that the robots can perform the front-end optimization locally (finding $(\tilde{R}_{ij}, \tilde{\tau}_{ij})$ for each edge $(i, j)$ in their trajectories), PGO addresses the back-end stage of SLAM.
    The objective function of PGO, in which the robots determine the set of poses (consisting of a rotation $R_i$ and translation $\tau_i$ for each pose $i$) that best explain the relative pose estimates $(\tilde{R}_{ij}, \tilde{\tau}_{ij})$, is separable and therefore amenable to distributed optimization techniques:}}
    \[
    \min_{\{(R_i, \tau_i)\}_{i = 1}^n}
    \sum_{(i,j)\in\mathcal{E}}\frac{\omega_{ij}}{2}\norm{R_j - R_i \tilde{R}_{ij} }^2_F + \frac{w_{ij}}{2}\norm{\tau_j - \tau_i - R_i \tilde{\tau}_{ij}}^2_2
    \]
   
    {{While PGO specifically addresses solving the back-end of SLAM}}, some existing distributed techniques that do not rely on distributed optimization have also been proposed for the front-end, e.g., \cite{cieslewski2017efficient}. We refer to \cite{durrant2006simultaneous, bailey2006simultaneous, grisetti2010tutorial, ahmad2013cooperative} for additional details on SLAM and multi-robot SLAM.
    
    Distributed optimization algorithms can be readily applied to the graph-based SLAM problem in \eqref{eq:pose_graph_slam_problem_reformulated}. Moreover, we note that a number of related robotics problems --- including rotation averaging/synchronization and shape registration/alignment --- can be similarly reformulated into a separable form and subsequently solved using distributed optimization algorithms \cite{dang2016decentralized, alwan2015distributed, todescato2015distributed, tron2014distributed, sarlette2009consensus, oh2018distributed}.

    \subsection{Multi-Robot Target Tracking}
    In the multi-robot target tracking problem, a group of robots collect measurements of an agent of interest (referred to as a target) and seek to collectively estimate the trajectory of the target. Multi-robot target tracking problems arise in many robotics applications ranging from environmental monitoring and surveillance to autonomous robotics applications such as autonomous driving, where the estimated trajectory of the target can be leveraged for scene prediction to enable safe operation. Figure \ref{fig:BLE_schematic} provides an illustration of the multi-robot target tracking problem where a group of four quadrotors make noisy observations of the flagged ground vehicle (the target). Each colored cone represents the region where each quadrotor can observe the vehicle, given the limited measurement range of the sensors onboard the quadrotor.
		
	\begin{figure}[tb]
        \centering
        \includegraphics[width=0.7\columnwidth]{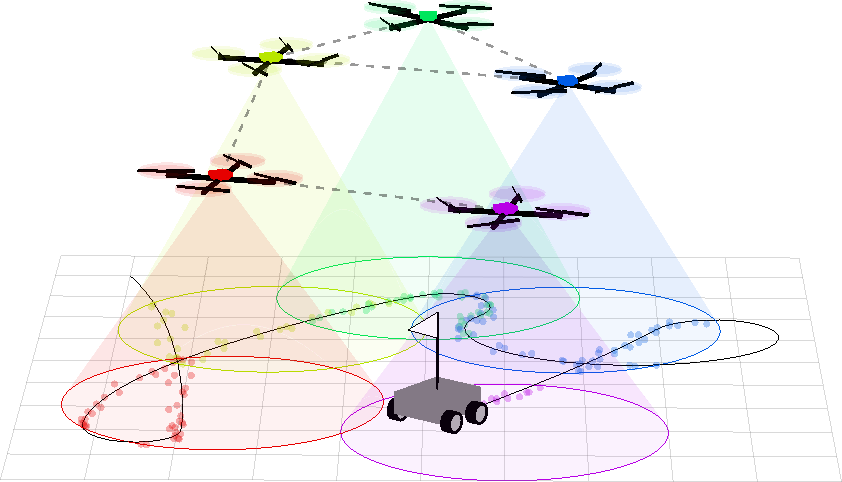}
        \caption{A multi-robot target tracking scenario, with four quadrotors (the robots) making noisy observations of the flagged ground vehicle (the target). The colored cones represent the regions where each quadrotor can observe the vehicle, given the limited measurement range of the sensors onboard each quadrotor.}
        \label{fig:BLE_schematic}
    \end{figure}
    
    Multi-robot target tracking problems can be posed as maximum a posterior (MAP) optimization problems where the robots seek to compute an estimate that maximizes the posterior distribution of the target's trajectory given the set of all observations of the target made by the robots. When a model of the dynamics of target is available, denoted by ${g: \mathbb{R}^{n} \rightarrow \mathbb{R}^{n}}$, the resulting optimization problem takes the form
    \begin{equation}
        \label{eq:target_tracking_problem}
        \begin{aligned}
            \underset{x}{\mathrm{minimize}} &\sum_{t=0}^{T - 1} \norm{x_{t + 1} - g(x_{t})}_{\Omega_{t}}^{2} \\
            &+ \sum_{i = 1}^{N} \sum_{t=0}^{T - 1} \norm{y_{i,t} - h_{i}(x_{t})}_{\Lambda_{i,t}}^{2},
        \end{aligned}
    \end{equation}
    where ${x_{t} \in \mathbb{R}^{n}}$ denotes the pose of the target at time $t$ and ${y_{i,t} \in \mathbb{R}^{m}}$ denotes robot $i$'s observation of the target at time $t$, over a duration of ${T + 1}$ timesteps. We represent the trajectory of the target with ${x = \left[x_{0}^{\top},x_{1}^{\top},\cdots,x_{T}^{\top}\right]^{\top}}$. While the first term in the objective function corresponds to the error between the estimated state of the target at a subsequent timestep and its expected state based on a model of its dynamics, the second term corresponds to the error between the observations collected by each robot and the expected measurement computed from the estimated state of the target, where the function ${h_{i}: \mathbb{R}^{n} \rightarrow \mathbb{R}^{m}}$ denotes the measurement model of robot $i$. Further, the information matrices ${\Omega_{t} \in \mathbb{R}^{n \times n}}$ and ${\Lambda_{i,t} \in \mathbb{R}^{m \times m}}$ for the dynamics and measurement models, respectively, weight the contribution of each term in the objective function appropriately, reflecting prior confidence in the dynamics and measurement models. The MAP optimization problem in \eqref{eq:target_tracking_problem} is not separable, hence, not amenable to distributed optimization, in its current form, due to coupling in the objective function arising from $x$. Nonetheless, we can arrive at a separable optimization problem through a fairly straightforward reformulation \cite{ola2020targettracking, shorinwa2023distributedtarget}. We can assign a local copy of $x$ to each robot, with $\hat{x}_{i}$ denoting robot $i$'s local copy of $x$. The reformulated problem becomes
    \begin{equation}
        \label{eq:target_tracking_problem_reformulated}
        \begin{aligned}
            \underset{\hat{x}}{\mathrm{minimize}} &\sum_{i = 1}^{N} \sum_{t=0}^{T - 1} \frac{1}{N}\norm{\hat{x}_{i,t + 1} - g(\hat{x}_{i,t})}_{\Omega_{t}}^{2} \\
            &+ \sum_{i = 1}^{N} \sum_{t=0}^{T - 1} \norm{y_{i,t} - h_{i}(\hat{x}_{i,t})}_{\Lambda_{i,t}}^{2} \\
            \mathrm{subject\ to}\ &\hat{x}_{i} = \hat{x}_{j} \quad \forall (i,j) \in \mathcal{E},
        \end{aligned}
    \end{equation}
    where ${\hat{x} = \left[\hat{x}_{1}^{\top},\hat{x}_{2}^{\top},\cdots,\hat{x}_{N}^{\top}\right]^{\top}}$. Following this reformulation, distributed optimization algorithms can be applied to compute an estimate of the trajectory of the target from \eqref{eq:target_tracking_problem_reformulated}.

    \subsection{Multi-Robot Task Assignment}
    In the multi-robot task assignment problem, we seek an optimal assignment of $N$ robots to $M$ tasks such that the total cost incurred in completing the specified tasks is minimized. However, we note that many task assignment problems consist of an equal number of tasks and robots. The standard task assignment problem has been studied extensively and is typically solved using the Hungarian method \cite{kuhn1955hungarian}. However, optimization-based methods have emerged as a competitive approach due to their amenability to task assignment problems with a diverse set of additional constraints, encoding individual preferences or other relevant problem information, making them a general-purpose approach. 
    
    The task assignment problem can be represented as a weighted bipartite graph: a graph whose vertices can be divide into two sets where no two nodes within a given set share an edge. Further, each edge in the graph has an associated weight. In task assignment problems, the edge weight $c_{i,j}$ represents the cost of assigning robot $i$ to task $j$. Figure \ref{fig:task_assign} depicts a task assignment problem represented by a weighted bipartite graph, with three robots and three tasks. Each robot knows its task preferences only and does not know the task preferences of other robots. Equivalently, the task assignment problem can be formulated as an integer optimization problem. Many optimization-based methods solve a  relaxation of the integer optimization problem. Generally, in problems with linear objective functions and affine constraints, these optimization-based methods are guaranteed to yield an optimal task assignment. The associated relaxed optimization problem takes the form
    \begin{equation}
        \label{eq:task_assignment_problem}
        \begin{aligned}
            \underset{x}{\mathrm{minimize}} &\sum_{i=1}^{N} c_{i}^{\top}x_{i} \\
            \mathrm{subject\ to}\ &\sum_{i = 1}^{N} x_{i} = 1_{M} \\
            &1_{M}^{\top}x_{i} = 1 \\
            & 0 \leq x \leq 1,
        \end{aligned}
    \end{equation}
    where ${x_{i} \in \mathbb{R}^{M}}$ denotes the optimization variable of robot $i$, representing its task assignment and ${x = \left[x_{1},x_{2},\cdots,x_{N}\right]}$. Although the objective function of \eqref{eq:task_assignment_problem} is separable, the optimization problem is not separable due to coupling of the optimization variables arising in the first constraint. We can obtain a separable problem, amenable to distributed optimization, by assigning a local copy of $x$ to each robot, resulting in the problem
    \begin{equation}
        \label{eq:task_assignment_problem_reformulated_v1}
        \begin{aligned}
            \underset{\hat{x}}{\mathrm{minimize}} &\sum_{i=1}^{N} c_{i}^{\top}\hat{x}_{i,i} \\
            \mathrm{subject\ to}\ &\sum_{i = 1}^{N} \hat{x}_{i,i} = 1_{M} \\
            &1_{M}^{\top}\hat{x}_{i,i} = 1 \\
            & 0 \leq \hat{x}_{i} \leq 1 \quad \forall i \in \mathcal{V} \\ 
            &\hat{x}_{i} = \hat{x}_{j} \quad \forall (i,j) \in \mathcal{E}
        \end{aligned}
    \end{equation}
    where ${\hat{x}_{i} \in \mathbb{R}^{M \times N}}$ denotes robot $i$'s local copy of $x$ and ${\hat{x} = \left[\hat{x}_{0},\hat{x}_{1},\cdots,\hat{x}_{N}\right]}$. Although the reformulation in \eqref{eq:task_assignment_problem_reformulated_v1} is simple, it does not scale efficiently with the number of robots and tasks. A more efficient reformulation can be obtained by considering the dual formulation of the task assignment problem. For brevity, we omit a discussion of this approach in this paper and refer readers to \cite{haksar2019consensus, liu2013optimal, giordani2010distributed, shorinwa2023distributedmultirobottask} where this reformulation scheme is discussed in detail.

	\begin{figure}[tb]
        \centering
        \includegraphics[width=0.7\columnwidth]{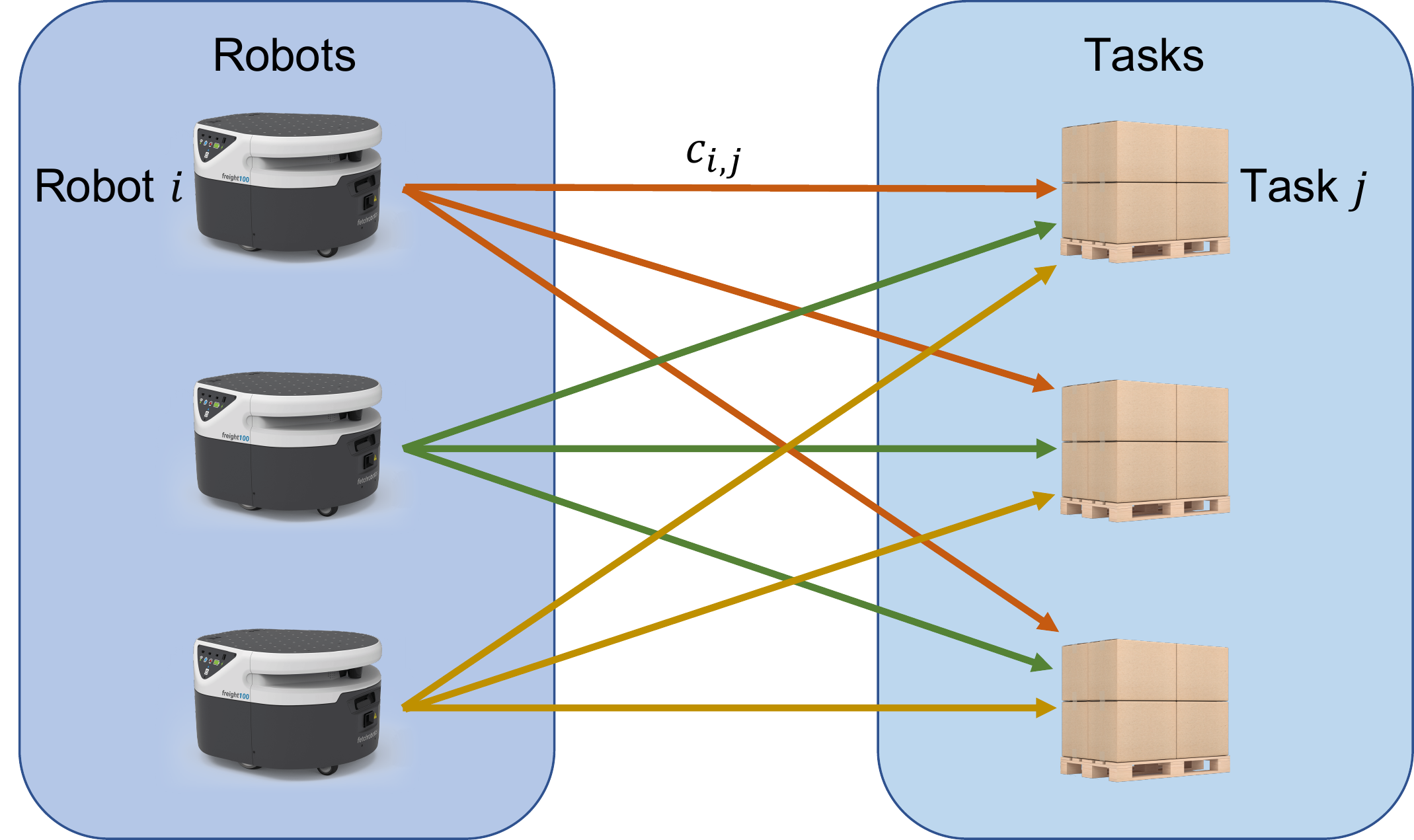}
        \caption{A multi-robot task assignment problem represented as a bipartite graph, with three (Fetch) robots and three tasks. An edge with weight $c_{i,j}$ between robot $i$ and task $j$ signifies the cost incurred by robot $i$ if it performs task $j$. In many problems, each robot's task preferences (edge weights) is neither known by other robots nor accessible to these robots.}
        \label{fig:task_assign}
    \end{figure}
    
    \subsection{Collaborative Planning, Control, and Manipulation}
    Generally, in collaborative planning problems, we seek to compute state and control input trajectories that enable a group of robots to reach a desired state configuration from a specified initial state, while minimizing a trajectory cost and without colliding with other agents. The related multi-robot control problem involves computing a sequence of control inputs that enables a group of robots to track a desired reference trajectory or achieve some specified task such as manipulating an object collaboratively.  Figure \ref{fig:collaborative_planning} shows a collaborative manipulation problem where three quadrotors move an object collaboratively. The dashed-line represents the reference trajectory for manipulating the load.
    	
	\begin{figure}[tb]
        \centering
        \includegraphics[width=0.7\columnwidth]{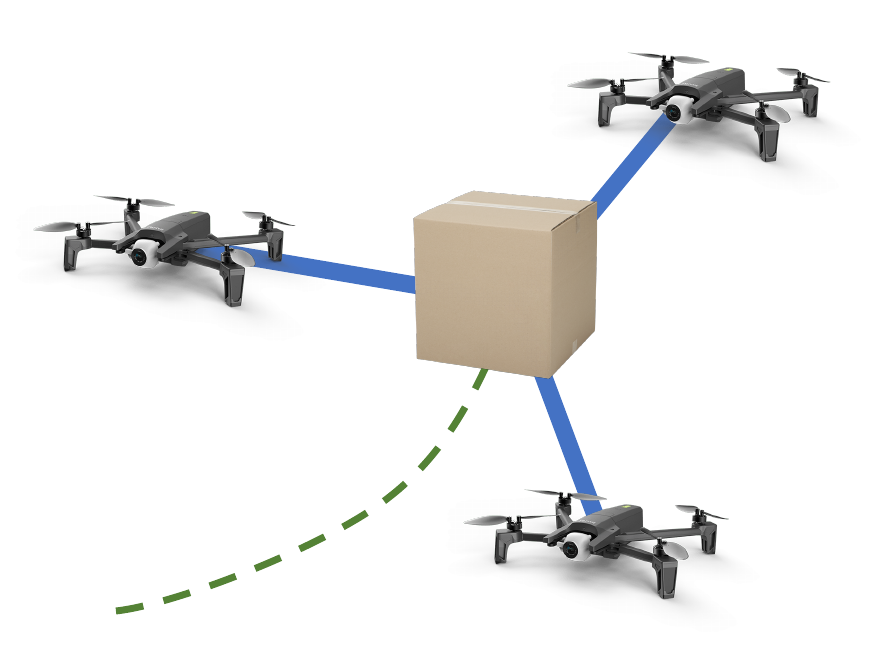}
        \caption{A multi-robot manipulation problem, with three quadrotors collaboratively manipulating a load rigidly attached to each quadrotor. The dashed-line represents the reference trajectory for manipulating the load.}
        \label{fig:collaborative_planning}
    \end{figure}

    Collaborative multi-robot planning, control, and manipulation problems have been well-studied, with a broad variety of methods devised for these problems. Among these methods, receding horizon or model predictive control (MPC) approaches have received notable attention due to their flexibility in encoding complex problem constraint and objectives.  In MPC approaches, these multi-robot problems are formulated as optimization problems over a finite time duration at each timestep. The resulting optimization problem is solved to obtain a sequence of control inputs over the specified time duration; however, only the initial control input is applied by each robot at the current timestep. At the next timestep, a new optimization problem is formulated, from which a new sequence of control inputs is computed to obtain a new control input for that timestep. This process is repeated until completion of the task. At time $t$, the associated MPC optimization problem has the form
    \begin{equation}
        \label{eq:multi_robot_mpc}
        \begin{aligned}
            \underset{x, u}{\mathrm{minimize}} &\sum_{i=1}^{N} f_{i}(x, u) \\
            \mathrm{subject\ to}\ &g(x,u) = 0 \\
            &h(x,u) \leq 0 \\
            & x_{i,0} = \bar{x}_{i} \quad \forall i \in \mathcal{V}
        \end{aligned}
    \end{equation}
    where ${x_{i} \in \mathbb{R}^{n_{i}}}$ denotes robot $i$'s state trajectory, ${u_{i} \in \mathbb{R}^{m_{i}}}$ denotes its control input trajectory, and ${x = \left[x_{1}^{\top},x_{2}^{\top},\cdots,x_{N}^{\top}\right]^{\top}}$ with ${u = \left[u_{1}^{\top},u_{2}^{\top},\cdots,u_{N}^{\top}\right]^{\top}}$. The objective function of robot $i$, ${f_{i}: \mathbb{R}^{\bar{n}} \times \mathbb{R}^{\bar{m}} \rightarrow \mathbb{R}}$, is often quadratic, given by
    \begin{equation}
        \label{eq:multi_robot_mpc_objective}
        \begin{aligned}
            f_{i}(x, u) =  (x_{i} - \tilde{x}_i)^{\top}&Q_{i}(x_{i} - \tilde{x}_i) \\
            & + (u_{i} - \tilde{u}_i)^{\top}R_i(u_{i} - \tilde{u}_i),
        \end{aligned}
    \end{equation}
    where $\tilde{x}_{i}$ and $\tilde{u}_{i}$ denote the reference state and control input trajectory, respectively, ${Q_{i} \in \mathbb{R}^{n_{i} \times n_{i}}}$ and ${R_{i} \in \mathbb{R}^{m_{i} \times m_{i}}}$ denote the associated weight matrices for the terms in the objective function, ${\bar{n} = \sum_{i = 1}^{N} n_{i}}$, and ${\bar{m} = \sum_{i = 1}^{N} m_{i}}$. The dynamics function of the robots is encoded in ${g: \mathbb{R}^{\bar{n}} \times \mathbb{R}^{\bar{m}} \rightarrow \mathbb{R}^{\bar{n}}}$. Further, other equality constraints can be encoded in $g$. Inequality constraints, such as collision-avoidance constraints and other state or control input feasibility constraints, are encoded in ${h: \mathbb{R}^{\bar{n}} \times \mathbb{R}^{\bar{m}} \rightarrow \mathbb{R}^{l}}$. In addition, the first state variable of each agent is constrained to be equal to its initial state, denoted by $\bar{x}_{i}$. In each instance of the MPC optimization problem, the initial state $\bar{x}_{i}$ of robot $i$ is specified as its current state at that timestep. Note that the MPC optimization problem in \eqref{eq:multi_robot_mpc} is not generally separable, depending on the equality and inequality constraints. However, a separable form of the problem can always be obtained by introducing local copies of the optimization variables that are coupled in \eqref{eq:multi_robot_mpc}. The functions $g$ and $h$ can also encode complementarity constraints for manipulation and locomotion problems that involve making and breaking rigid body contact \cite{shorinwa2021distributed}. In the extreme case, where the optimization variables are coupled in the objective function and equality and inequality constraints in \eqref{eq:multi_robot_mpc}, a suitable reformulation takes the form
    \begin{equation}
        \label{eq:multi_robot_mpc_reformulated}
        \begin{aligned}
            \underset{\hat{x}, \hat{u}}{\mathrm{minimize}} &\sum_{i=1}^{N} f_{i}(\hat{x}_{i}, \hat{u}_{i}) \\
            \mathrm{subject\ to}\ &g(\hat{x}_{i}, \hat{u}_{i}) = 0 \quad \forall i \in \mathcal{V} \\
            &h(\hat{x}_{i}, \hat{u}_{i}) \leq 0  \quad \forall i \in \mathcal{V}\\
            & \phi_{i}(\hat{x}_{i}) = \bar{x}_{i} \quad \forall i \in \mathcal{V} \\ 
            &\hat{x}_{i} = \hat{x}_{j} \quad \forall (i,j) \in \mathcal{E},
        \end{aligned}
    \end{equation}
    where the function $\phi_{i}$ outputs the first state variable corresponding to robot $i$, given the input $\hat{x}_{i}$, which denotes robot $i$'s local copy of $x$. Similarly, $\hat{u}_{i}$ denotes robot $i$'s local copy of $u$, with ${\hat{x} = \left[\hat{x}_{1}^{\top},\hat{x}_{2}^{\top},\cdots,\hat{x}_{N}^{\top}\right]^{\top}}$ and ${\hat{u} = \left[\hat{u}_{1}^{\top},\hat{u}_{2}^{\top},\cdots,\hat{u}_{N}^{\top}\right]^{\top}}$. Distributed optimization algorithms \cite{bento2013message, ferranti2018coordination, ola2020collab, shorinwa2023distributedmodelpredictive} can be employed to solve the resulting MPC optimization problem in \eqref{eq:multi_robot_mpc_reformulated}.

    \subsection{Multi-Robot Learning}
	Multi-robot learning entails the application of deep learning methods to approximate functions from data to solve multi-robot tasks, such as object detection, visual place recognition, monocular depth estimation, 3D mapping, and multi-robot reinforcement learning. Consider a general multi-robot supervised learning problem where we aim to minimize a loss function over labeled data collected by all the robots.  We can write this as
	\[
	\min_{\theta}\sum_{i = 1}^N\sum_{(x_{ij}, y_{ij})\in D_i}l(y_i, f(x_i; \theta)),
	\]
	where $l(\cdot,\cdot)$ is the loss function, $(x_{ij}, y_{ij})$ is data point $j$ collected by robot $i$ with feature vector $x_{ij}$ and label $y_{ij}$, $D_i$ is the set of data collected by robot $i$, $\theta$ are the neural network weights, and $f(x;\theta)$ is the neural network parameterized function we desire to learn.  By creating local copies of the neural network weights $\theta_i$ and adding consensus constraints $\theta_i = \theta_j$, we can put problem in the form (\ref{eq:general_problem_with_agreement}), so it is amenable to distributed optimization. We stress that this problem encompasses a large majority of problems in supervised learning.  See \cite{yu2022dinno} for an ADMM-based distributed optimization approach to solving this problem.	  
	
	Beyond supervised learning, many multi-robot learning problems are formulated within the framework of reinforcement learning. In these problems, the robots learn a control policy by interacting with their environments by making sequential decisions. The underlying control policy, which drives these sequential decisions, is iteratively updated to optimize the performance of all agents on a specified objective using the information gathered by each robot during its interaction with its environment. Figure \ref{fig:multi_robot_learning} illustrates the reinforcement learning paradigm, where a group of robots learn from experience. Each robot takes an action and receives an observation (and a reward), which provides information on the performance of its current control policy in achieving its specified objective.
	
	\begin{figure}[tb]
        \centering
        \includegraphics[width=0.9\columnwidth]{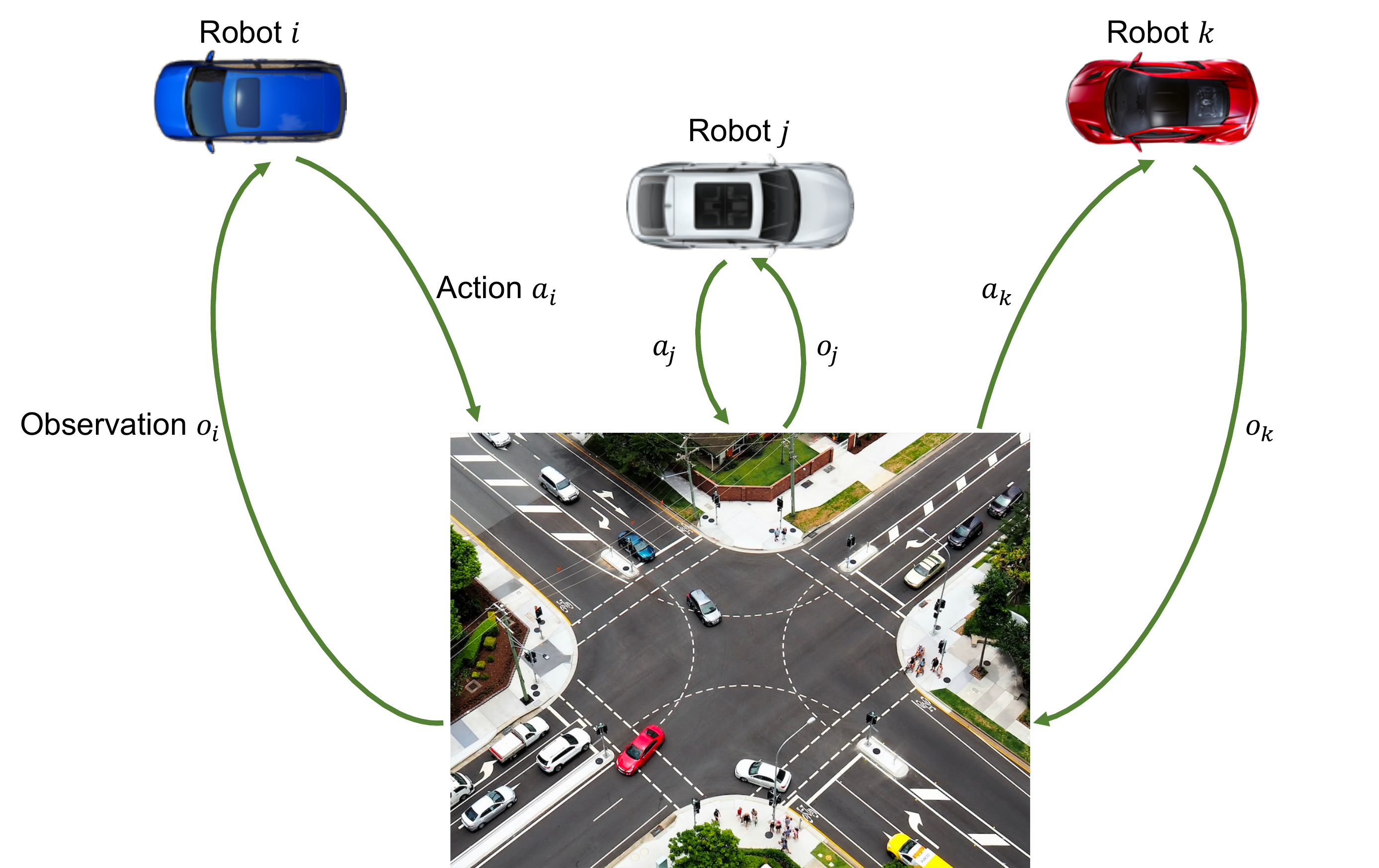}
        \caption{In multi-robot reinforcement learning problems, a group of robots compute a control policy from experience by making sequential decisions while interacting with their environment. Each robot takes an action and receives an observation (and a reward), which provides information on its performance in accomplishing a specified task.}
        \label{fig:multi_robot_learning}
    \end{figure}
	
	Reinforcement learning approaches can be broadly categorized into value-based methods and policy-based methods. Value-based methods seek to compute an estimate of the optimal action-value function --- the $Q$-function --- which represents the expected discounted reward starting from a given state and taking a given action. An optimal policy can be extracted from the estimated $Q$-function by selecting the action that maximizes the value of the $Q$-function at a specified state. In deep value-based methods, deep neural networks are utilized in approximating the $Q$-function. In contrast, policy-based methods seek to find an optimal policy by directly searching over the space of policies. In deep policy-based methods, the control policy is parameterized using deep neural networks. In general, the agents seek to maximize the expected infinite-horizon discounted cumulative reward, which is posed as the optimization problem
    \begin{equation}
    	\label{eq:multi_robot_learning_prob}
    	\begin{aligned}
    		\underset{\theta}{\mathrm{maximize}}\ &\mathbb{E}_{\pi_{\theta}} \left[ \sum_{t \geq 0} \gamma^{t} \sum_{i = 1}^{N} R_{i}(s_{i, t}, a_{i,t}) \mid s_{i,0} = \bar{s}_{i}\right],
    	\end{aligned}
    \end{equation}
    where $\pi_{\theta}$ denotes the control policy parameterized by $\theta$, ${\gamma \in \mathbb{R}}$ denotes the discount factor (${\gamma \in (0, 1)}$), $s_{i,t}$ denotes the state of robot $i$ at time $t$, $a_{i,t}$ denotes its action at time $t$, $\bar{s}_{i}$ denotes its initial state, ${R_{i}: \mathcal{S}_{i} \times \mathcal{A}_{i} \rightarrow \mathbb{R}}$ denotes the reward function of robot $i$, and $N$ denotes the number of robots. The optimization problem in \eqref{eq:multi_robot_learning_prob} is not separable in its current form. However, due to the linearity of the expectation operator, the optimization problem in \eqref{eq:multi_robot_learning_prob} can be equivalently expressed as
    \begin{equation}
    	\label{eq:multi_robot_learning_prob_reformulated}
    	\begin{aligned}
    		\underset{\hat{\theta}_{1},\cdots,\hat{\theta}_{N}}{\mathrm{maximize}} &\sum_{i = 1}^{N} \mathbb{E}_{\pi_{\hat{\theta}_{i}}} \left[ \sum_{t \geq 0} \gamma^{t} R_{i}(s_{i, t}, a_{i,t}) \mid s_{i,0} = \bar{s}_{i}\right] \\
    		\mathrm{subject\ to}\ &\hat{\theta}_{i} = \hat{\theta}_{j} \quad \forall (i,j) \in \mathcal{E},
    	\end{aligned}
    \end{equation}
    which is separable among the $N$ robots. Hence, the resulting problem can be readily solved using distributed optimization algorithms for reinforcement learning problems, such as distributed $Q$-learning and distributed actor-critic methods \cite{zhang2018fully, zhang2019distributed, oroojlooyjadid2019review}.
   
  {
     \section{Notes on Implementation, Practical Performance, and Limitations}
   \label{sec:practical_notes}
   Here, we highlight some relevant issues that arise in the application of distributed optimization algorithms in robotics problems. In Table~\ref{tab:algorithm_summary}, we highlight a few characteristics of the algorithms in each class of distributed optimization problems. We note that the properties of each algorithm class displayed in Table \ref{tab:algorithm_summary} are based on the representative algorithm considered in the algorithm class. We emphasize that subsequent research efforts have been devoted to the derivation of algorithms that address the practical issues faced by many of the existing algorithms. In this section, we describe alternative distributed algorithms that address these issues, often at the expense of convergence speed.

    \subsection{Selection of a Stochastic Matrix}
	Distributed first-order algorithms and distributed sequential convex programming algorithms require the specification of a stochastic matrix, which must be compatible with the underlying communication network. 
    In general, generating compatible row-stochastic and column-stochastic matrices for directed communication networks does not pose a significant challenge. To obtain a row-stochastic matrix, each robot assigns a weight to all its in-neighbors such that the sum of all its weights equals one. Similarly, to obtain a column-stochastic matrix, each robot assigns a weight to all its out-neighbors such that the sum of all its weights equals one. In contrast, generating doubly-stochastic matrices for directed communication networks is nontrivial if each robot does not know the global network topology. Consequently, in general, algorithms which require doubly-stochastic matrices are unsuitable for problems with directed communication networks. 
	
	A number of distributed first-order algorithms allow for the specification of row-stochastic or column-stochastic matrices, making this class of algorithms appropriate for problems with directed communication networks, unlike distributed sequential convex programming algorithms, which generally require the specification of a doubly-stochastic weighting matrix. Furthermore, a number of distributed sequential convex programming algorithms require symmetry of the doubly-stochastic weighting matrix \cite{Mokhtari2015, Mokhtari2016, Eisen2019, mansoori2019fast}, posing an even greater challenge in problems with directed networks.
	
	The specific choice of a doubly-stochastic weighing matrix may vary depending on the assumptions made on what global knowledge is available to the robots on the network.  The problem of choosing an optimal weight matrix is discussed thoroughly in \cite{xiao2004fast}, in which the authors show that achieving the fastest possible consensus can be posed as a semidefinite program, which a computer with global knowledge of the network can solve efficiently. However, we cannot always assume that global knowledge of the network is available, especially in the case of a time-varying topology. In most cases, Metropolis weights facilitate fast mixing without requiring global knowledge, with the assumption that the communication network is undirected with bi-directional communication links. Each robot can generate its own weight vector after a single communication round with its neighbors. In fact, Metropolis weights perform only slightly sub-optimally compared to centralized optimization-based methods \cite{jafarizadeh2010weight}:
    	\begin{equation}
    	w_{ij} = \begin{cases}
    	        \frac{1}{\max\{\vert \mathcal{N}_i\vert, \vert \mathcal{N}_j \vert\} }& j \in \mathcal{N}_i, \\
    	        1- \sum_{j^\prime \in \mathcal{N}_i} w_{ij^\prime} & i = j,
    	        \\
    	        0 & \text{else}.
    	\end{cases}
    	\end{equation}
    	
     Distributed algorithms based on ADMM do not require the specification of a stochastic weighting matrix. However, C-ADMM \cite{mateos2010distributed} and other distributed variants assume that the communication network between all robots is bi-directional, which makes these algorithms unsuitable for problems with directed communication networks. A number of distributed ADMM algorithms for problems with directed communication networks have been developed \cite{khatana2020d, khatana2020dc, rokade2020distributed}. Owing to the absence of bi-directional communication links between the robots, these algorithms utilize a dynamic average consensus scheme to update the slack variables at each iteration, which merges information from a robot and its neighbors using a stochastic weighting matrix. However, some of these distributed algorithms require the specification of a doubly-stochastic weighting matrix \cite{rokade2020distributed}, which introduces notable challenges in problems with directed communication networks, while others allow for the specification of a column-stochastic weighting matrix \cite{khatana2020dc}.

	\subsection{Initialization}
	In general, distributed optimization algorithms allow for an arbitrary initialization of the initial solution of each robot, in convex problems. However, these algorithms often place stringent requirements on the initialization of the algorithms' parameters. DFO methods require initialization of the step-size and often place conditions on the value of the step-size to guarantee convergence. Some distributed gradient tracking algorithms \cite{nedic2017digging, qu2017harnessing} assume all robots use a common step-size, requiring coordination among all robots. Selecting a common step-size might involve the execution of a consensus procedure by all robots, with additional computation and communication overhead. In algorithms which utilize a fixed step-size, this procedure only needs to be executed once, at the beginning of the optimization algorithm. ADMM and its distributed variants require the selection of a common penalty parameter $\rho$. Consequently, all robots must coordinate among themselves in selecting a value for $\rho$, introducing some challenges, particularly in problems where the convergence rate depends strongly on the value of $\rho$. Initialization of these algorithm-specific parameters have a significant impact on the performance of each algorithm.

 In general, the performance of each distributed algorithm that we consider is sensitive to the choice of parameters especially when local objective functions are poorly conditioned.  For instance, in DFO methods, choosing $\alpha$ too large leads to divergence of the individual variables, while too small a value of $\alpha$ causes slow convergence. Similarly, C-ADMM (Algorithm \ref{alg:C-ADMM}) has a convergence rate that is highly sensitive to the choice of $\rho$, though convergence is guaranteed for all $\rho > 0$.  We study the sensitivity of the convergence rate to parameter choice in each simulation in Section \ref{sec:simulations}.  However, the optimal parameter choice for a particular example is not prescriptive for the tuning of other implementations.
 { The optimal step-size for a particular algorithm depends on many factors, including the network size, the network connectivity, and the underlying problem. For instance, the size of the network affects the value of the step-size that achieves optimal convergence, as well as the maximum rate of convergence itself.}
 Furthermore, while analytical results for optimal parameter selection are available for many of these algorithms, a practical parameter-tuning procedure is useful if an implementation does not exactly adhere to the assumptions in the literature.

 In the case that parameter tuning is essential to performance, it can be reasonable to select suitable parameters for an implementation before deploying a system, either using analytical results or simulation. The most general (centralized) procedure for parameter tuning involves comparing the convergence performance of the system on a known problem for different parameter values.  While a uniform sweep of the parameter space may be effective for small problems or parameter-insensitive methods, it is not computationally efficient.  Given the convergence rate of a distributed method at particular choices of parameter, \textit{bracketing} methods provide parameter selections to more efficiently find the convergence-rate-minimizing parameter. For instance, Golden Section Search (GSS) provides a versatile approach for tuning a scalar parameter \cite{press1989numerical}.
{ Finding the optimal step-size in one instance of a problem often provides reasonable parameter choices for a problem of similar size, connectivity, and structure.}

	\subsection{Dynamic or Lossy Communication}\label{subsection:dynamic-communication}
	In practical situations, the communication network between robots changes over time as the robots move, giving rise to a time-varying communication graph. Networked robots in the real world can also suffer from dropped message packets as well as failed hardware or software components. Lossy communication can be a result of both networks where there are many robots and communication signals interfere, and in situations where robot's have unstable communication links (e.g. wireless connections close to range limits). Generally, distributed first-order optimization algorithms are amenable to problems with dynamic communication networks and are guaranteed to converge to the optimal solution provided that the communication graph is $B$-connected for undirected communication graphs or $B$-strongly connected for directed communication graphs \cite{nedic2017digging}, which implies that the union of the communication graphs over $B$ consecutive time-steps is connected or strongly-connected respectively. This property is also referred to as bounded connectivity. This assumption ensures the diffusion of information among all robots. Unlike DFO algorithms, many distributed sequential convex programming algorithms assume the communication network remains static. Nevertheless, a few distributed sequential programming algorithms are amenable to problems with dynamic communication networks \cite{di2016next, sun2017distributed} and converge to the optimal solution of the problem under the assumption that the sequence of communication graphs is $B$-strongly connected. Some distributed ADMM algorithms are not amenable to problems with dynamic communication networks.  This is an interesting avenue for future research.

    Similarly, dropped messages or packets can be modeled as changes to edges in the communication graph where an edge temporarily becomes directed.  In modern mesh networking protocols, dropped packets can be detected through packet acknowledgement and the data can either be resent, or the robots can choose to ignore that communication link during the given iteration of distributed optimization.
    We explore the effect of dropping edges from the communication network in Sec.~\ref{sec:simulations}, Fig.~\ref{fig:dropped_edges_convergence}.

    \subsection{Synchronization}
    Synchronization, in the context of distributed optimization, is the assumption that robots compute their local updates and communicate at the same time, and ensures that each robot has up-to-date communicated variables from its neighbors. { Many distributed optimization algorithms require synchronous execution for guaranteed convergence to an optimal solution \cite{nedic2017digging, shi2015extra, di2016next, Mokhtari2015, mateos2010distributed, ola2020SOVA}.} In practice when networks are have many agents or heterogeneous computation capability, it is unlikely that all robots will finish their local computation/communication at exactly the same time, and therefore some practical synchronization scheme is required. Fortunately, one simple solution is to have each robot wait to receive updates from each of its neighbors before proceeding with its next iteration of distributed optimization. This is the decentralized version of a barrier algorithm \cite{arenstorf1989comparing} in parallel computing. When all robots require roughly the same amount of time to perform each iteration, this simple barrier approach has a negligible impact on the time to convergence of a distributed optimization algorithm. However, if some subset of the robots are much slower than the others then this barrier approach can result in long idle times for some of the robots, and longer time to convergence.

    Alternatively, DFO algorithms (DIGing, EXTRA, etc.) are generally fairly amenable to asynchronous execution, and some other methods are explicitly designed for asynchronous execution \cite{lian2018asynchronous}.

    }
  
 \section{Distributed Multi-drone Vehicle Tracking: A Case Study}
	\label{sec:simulations}
	
    { We illustrate the implementation of distributed optimization methods using a simulation of a multi-drone vehicle target tracking problem as a case study. We emphasize that the same principles apply to a broad class of robotics problems that we have outlined in Sec.~\ref{sec:multi_robot_optimization_problems}.}  In addition, we implement the distributed optimization algorithm C-ADMM on a network of Raspberry Pis communicating with XBee modules to demonstrate a distributed optimization algorithm on hardware.
	
	\subsection{Simulation Study}
	In this simulation, we consider a distributed multi-drone vehicle target tracking problem in which robots connected by a communication graph, $\mc{G} = (\mc{V}, \mc{E})$, each record range-limited linear measurements of a moving target, and seek to collectively estimate the target's entire trajectory. We assume that each drone can communicate locally with nearby drones over the undirected communication graph $\mc{G}$.   The drones all share a linear model of the target's dynamics as \begin{equation}
	    x_{t+1} = A_t x_t + w_t,
	\end{equation} 
	where $x_t \in \mathbb{R}^4$ represents the position and velocity of the target in some global frame at time $t$, $A_t$ is the dynamics matrix associated with a linear model of the target's dynamics, and $w_t \sim \mc{N}(0, Q_t)$ represents process noise (including the unknown control inputs to the target).  Restricting our case study to a linear target model in this tutorial ensures that the underlying optimization problem is convex, leading to strong convergence guarantees and robust numerical properties for our algorithm.  A more expressive nonlinear model can also be used, but this requires a more sophisticated distributed optimization algorithm with more challenging numerical properties. 
	At every time-step when the target is sufficiently close to a drone $i$ (which we denote by $t \in \mc{T}_i$), that robot collects an observation according to the linear measurement model
	\begin{equation}
	    y_{i,t} = C_{i,t} x_t + v_{i, t} \: ,
	\end{equation} where $y_{i,t} \in \mathbb{R}^2$ is a positional measurement, $C_{i,t}$ is the measurement matrix of drone $i$, and $v_{i, t} \sim \mathcal{N}(0, R_{i, t})$ is measurement noise.  We again assume a linear measurement model to keep this case study as simple as possible.  A nonlinear model can also be used.  
	
	All of the drones have the same model for the prior distribution of the initial state of the target $\mc{N}(\bar{x}_0, \bar{P}_0)$, where ${\bar{x}_0 \in \mathbb{R}^{4}}$ denotes the mean and ${\bar{P}_0 \in \mathbb{R}^{4 \times 4}}$ denotes the covariance.
	The global cost function is of the form 
	\begin{equation}
	    \label{eq:trajectory_estimation}
	    \begin{aligned}
	    f(x) = & \norm{x_{0} - \bar{x}_{0}}_{\bar{P}_{0}^{-1}}^{2} + \sum_{t=1}^{T-1} \norm{x_{t+1} - A_t x_{t}}_{Q_t^{-1}}^{2}  \\
	    & + \sum_{i \in \mc{V}} \sum_{t \in \mc{T}_i} \norm{y_{i,t} - C_{i, t} x_{t}}_{R^{-1}}^{2},
	    \end{aligned}
	\end{equation}
	while the local cost function for drone $i$ is
	\begin{equation}
	    \label{eq:trajectory_estimation_local}
	    \begin{aligned}
	    f_i(x) = & \frac{1}{N}\norm{x_{0} - \bar{x}_{0}}_{\bar{P}_{0}^{-1}}^{2} + \sum_{t=1}^{T-1} \frac{1}{N} \norm{x_{t+1} - A_t x_{t}}_{Q_t^{-1}}^{2}  \\
	    & + \sum_{t \in \mc{T}_i} \norm{y_{i,t} - C_{i, t} x_{t}}_{R^{-1}}^{2}.
	    \end{aligned}
	\end{equation}
	
	In our results, we consider only a batch solution to the problem (finding the full trajectory of the target given each robot's full set of measurements).  Methods for performing the estimate in real-time through filtering and smoothing steps have been well studied, both in the centralized and distributed case \cite{olfati2007distributed}. An extended version of this multi-robot tracking problem is solved with distributed optimization in \cite{ola2020targettracking}. A rendering of a representative instance of this multi-robot tracking problem is shown in Figure \ref{fig:BLE_schematic}.

    In Figures \ref{fig:BLE_cpi} and \ref{fig:BLE_step_sense}, several distributed optimization algorithms are compared on an instance of the distributed multi-drone vehicle tracking problem. For this problem instance, 10 simulated drones seek to estimate the target's trajectory over 16 time steps resulting in a decision variable dimension of $n = 64$. We compare four distributed optimization methods which we consider to be representative of the taxonomic classes outlined in the sections above: C-ADMM \cite{mateos2010distributed}, EXTRA \cite{shi2015extra}, DIGing \cite{nedic2017digging}, and NEXT-Q \cite{di2016next}. Figure \ref{fig:BLE_cpi} shows that C-ADMM and EXTRA have similar fast convergence rates per iteration while DIGing and NEXT-Q are 4 and 15 times slower respectively to converge below an MSE of $10^{-6}$. The step-size hyperparameters for each method are computed by Golden Section Search (GSS) (for NEXT-Q, which uses a two parameter decreasing step-size, we fix one according to the values recommended in \cite{di2016next}).
    
   We note that tuning is essential for achieving robust and efficient convergence with most distributed optimization algorithms. Figure \ref{fig:BLE_step_sense} shows the sensitivity of these methods to variation in step-size, and highlights that three of the methods (all except C-ADMM) {diverge for large step-sizes. In the case of EXTRA in this example, the optimal step-size is close in value to step-sizes that lead to divergence, posing a practical challenge for parameter tuning.} While C-ADMM seems to be the most effective algorithm in this problem instance, we note that other algorithms have properties that are advantageous in other instances of this problem or other problems. { Furthermore, the optimal step-size depends on the problem structure. For instance, in this problem, as the number of agents increase, the optimal step-size decreases for C-ADMM and increases for the other methods.}

	\begin{figure}[tb]
	    \centering
	    \includegraphics[width=\linewidth]{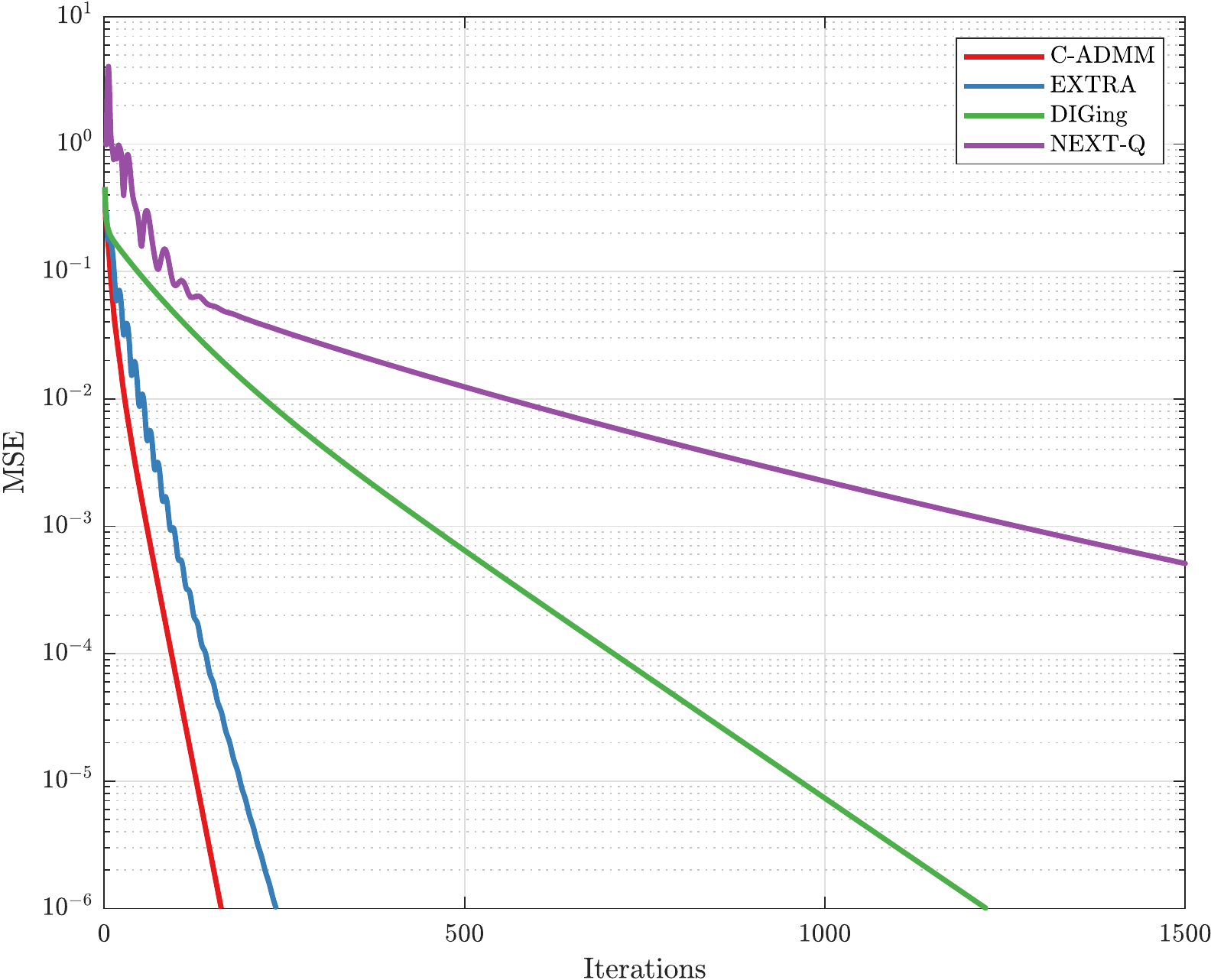}
	    \caption{Mean Square Error (MSE) per iteration on a distributed multi-drone vehicle target tracking problem with $N=10$ and $n = 64$. }
	    \label{fig:BLE_cpi}
	\end{figure}

	\begin{figure}[tb]
        \centering
        \includegraphics[width=\columnwidth]{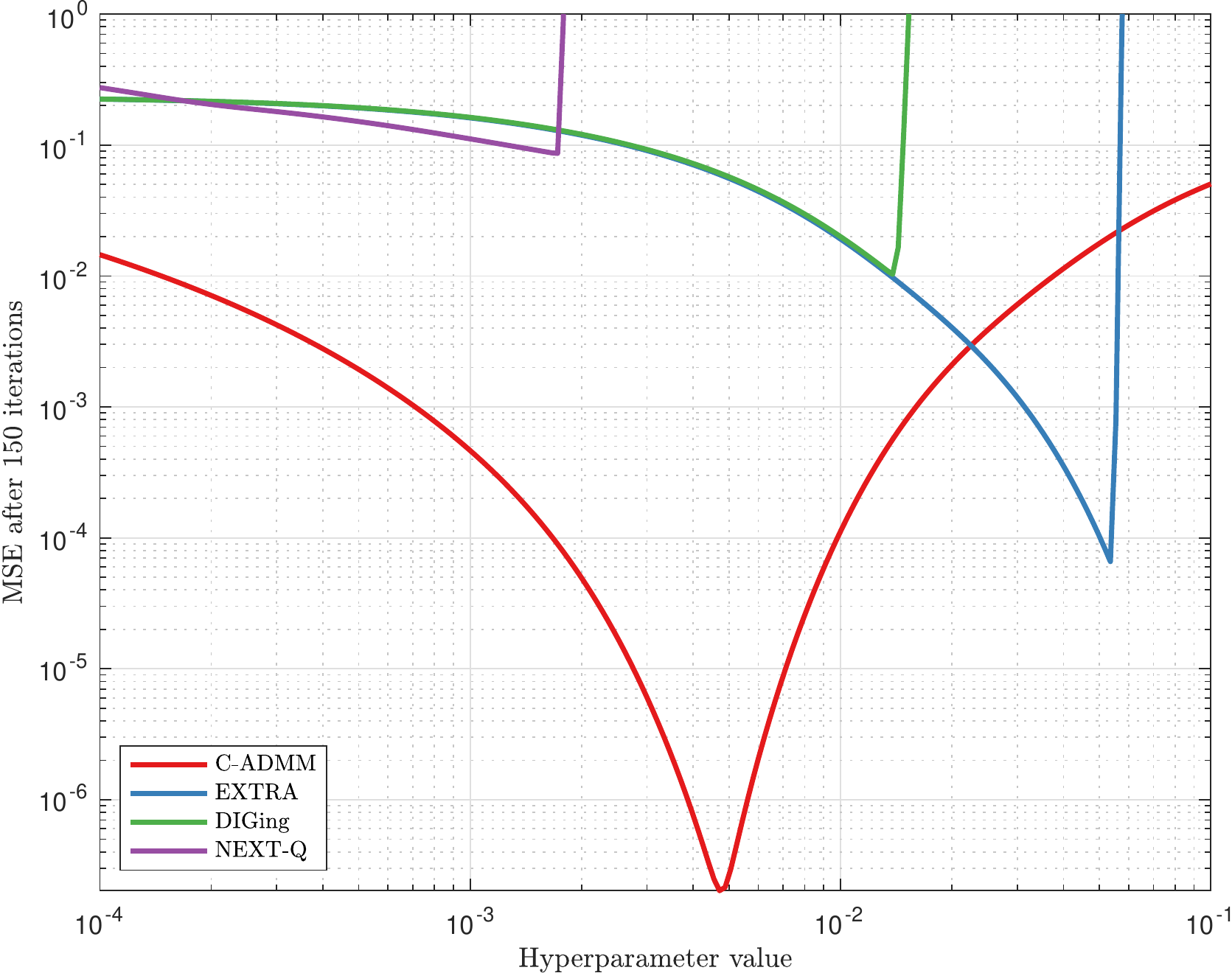}
        \caption{Hyperparameter sensitivity sweep for a distributed multi-drone vehicle target tracking problem with $N=20$ and $n = 64$. %
        EXTRA, DIGing, and NEXT-Q diverge when their respective step-sizes are too large, while C-ADMM converges over all choices of $\rho$. (C-ADMM  values are reported with respect to $\rho/100$ in order to fit on the same axes as the other methods.)
        }
        \label{fig:BLE_step_sense}
    \end{figure}

    \begin{figure}[tb]
        \centering
        \includegraphics[width=\columnwidth]{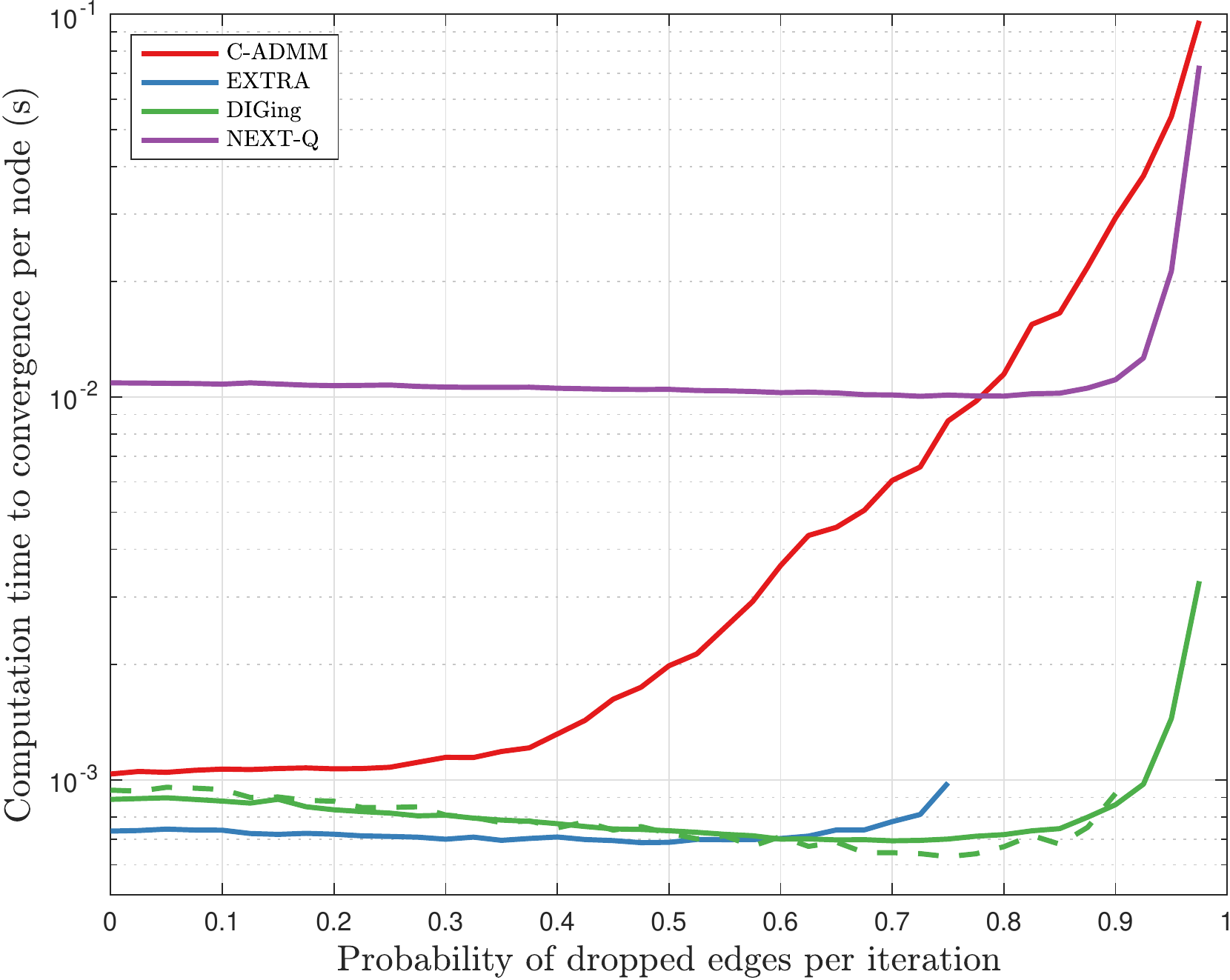}
        \caption{{Computation time to convergence as a function of the probability of dropped edges in a mesh network, averaged over 50 trials using a geometric random graph with $N = 20$. The stopping condition for each trial is a normalized MSE of $10^{-6}$. Each undirected edge is dropped with the given probability at every iteration. DIGing is the only method considered that can handle directional lost edges (dashed line). Implementations use optimal hyper-parameters, which vary according to the probability of dropped edges.}}
        \label{fig:dropped_edges_convergence}
    \end{figure}

    {
    As discussed in Section~\ref{subsection:dynamic-communication}, the convergence of distributed optimization algorithms may degrade under dynamic or lossy communication.  In Figure~\ref{fig:dropped_edges_convergence}, we demonstrate this effect given a geometric random graph with $N=20$. For all four methods considered, a low probability of missing edges does not significantly degrade convergence compared to a static network. In particular,  DIGing and NEXT-Q are robust to dropped edges, while EXTRA diverges for high rates of dropped edges and C-ADMM converges for carefully chosen values of $\rho$ but at orders of magnitude increased computation time. While C-ADMM converges in fewer iterations than the other methods in the examples of Figures~\ref{fig:BLE_cpi} and \ref{fig:BLE_step_sense}, the dynamic graph topology in Figure~\ref{fig:dropped_edges_convergence} means that we cannot precompute matrix inverses, resulting in slower computation per iteration (reported computation time is based on a MacBook Pro with M1 Pro chip and 16GB unified memory). Of the methods considered, only DIGing handles directed dropped edges.  While NEXT also addresses directed network communication, it requires a doubly-stochastic matrix at each iteration. Fast, distributed construction of doubly-stochastic matrices is still an open question \cite{xi2017distributed}.}

    \subsection{Hardware Implementation}
    In this section, we discuss our implementation of the \mbox{C-ADMM} algorithm on hardware. Each robot is equipped with local computational resources and communication hardware necessary for peer-to-peer communication with other neighboring robots. In the following discussion, we provide details of the hardware platform, the underlying communication network between robots, and the optimization problem considered in this section. 
    
    We consider the linear least-squares optimization problem
    \begin{equation}
        \min_p \sum_{i=1}^N (G_i p - z_i)^\top M_i (G_i p - z_i), \label{eqn:factored_dist_eqn}
    \end{equation}
    with the optimization variable ${p \in \mathbb{R}^{32}}$, ${G_{i} \in \mathbb{R}^{m_{i} \times 32}}$, ${M_{i} \in \mathbb{R}^{m_{i} \times m_{i}}}$, ${z_{i} \in \mathbb{R}^{m_{i}}}$, and ${N = 3}$ robots, where $m_{i}$ depends on the number of measurements available to robot $i$. In this experiment, we have ${m_{1} = 3268}$, ${m_{2} = 5422}$, and ${m_{3} = 3528}$. We implement C-ADMM to solve the problem, with a state size consisting of $32$ floating-point variables.  
    
    The core communication infrastructure that we use are Digi XBee DigiMesh 2.4 radio frequency mesh networking modules which allow for peer-to-peer communication between robots. Local computation for each robot is performed using Raspberry Pi 4B single board computers. The lower level mesh network is managed by the DigiMesh software, and we interact with it through XBee Python Library.

    We utilize the neighbor discovery Application Programming Interface (API) provided by Digi International to enable each robot to identify other neighboring robots. This approach resulted in a fully-connected communication network, considering the XBee radios have an indoor range of up to $90$m and an outdoor range of up to $1500$m. The XBee modules used in our experiments have a maximum payload size of $92$ bytes. However, the local variable of each robot in our experiment consists of $32$ floating-point variables, which exceeds the maximum payload size that can be transmitted by the XBee radios at each broadcast round, presenting a communication challenge. To overcome this challenge, we break up the local variables into a series of packets of size $92$ bytes and perform multiple broadcast rounds. { The resulting implementation required approximately 5.5 sec per round of} {
    communication in C-ADMM (i.e. for all the robots to exchange their decision variable information). In contrast the Raspberry Pi computation for each iteration of C-ADMM was approximately 15 microseconds, so communication time was approximately 5 orders of magnitude slower than computation time in our implementation. This slow communication speed is due to the severe bandwidth limitations of the XBee radios. We expect an optimized implementation over a state-of-the art 5 Gbit/sec WiFi or 5G network would reduce this communication time to about $0.2$ microseconds per round.} 

    	\begin{figure}[tb]
        \centering
        \includegraphics[width=\columnwidth]{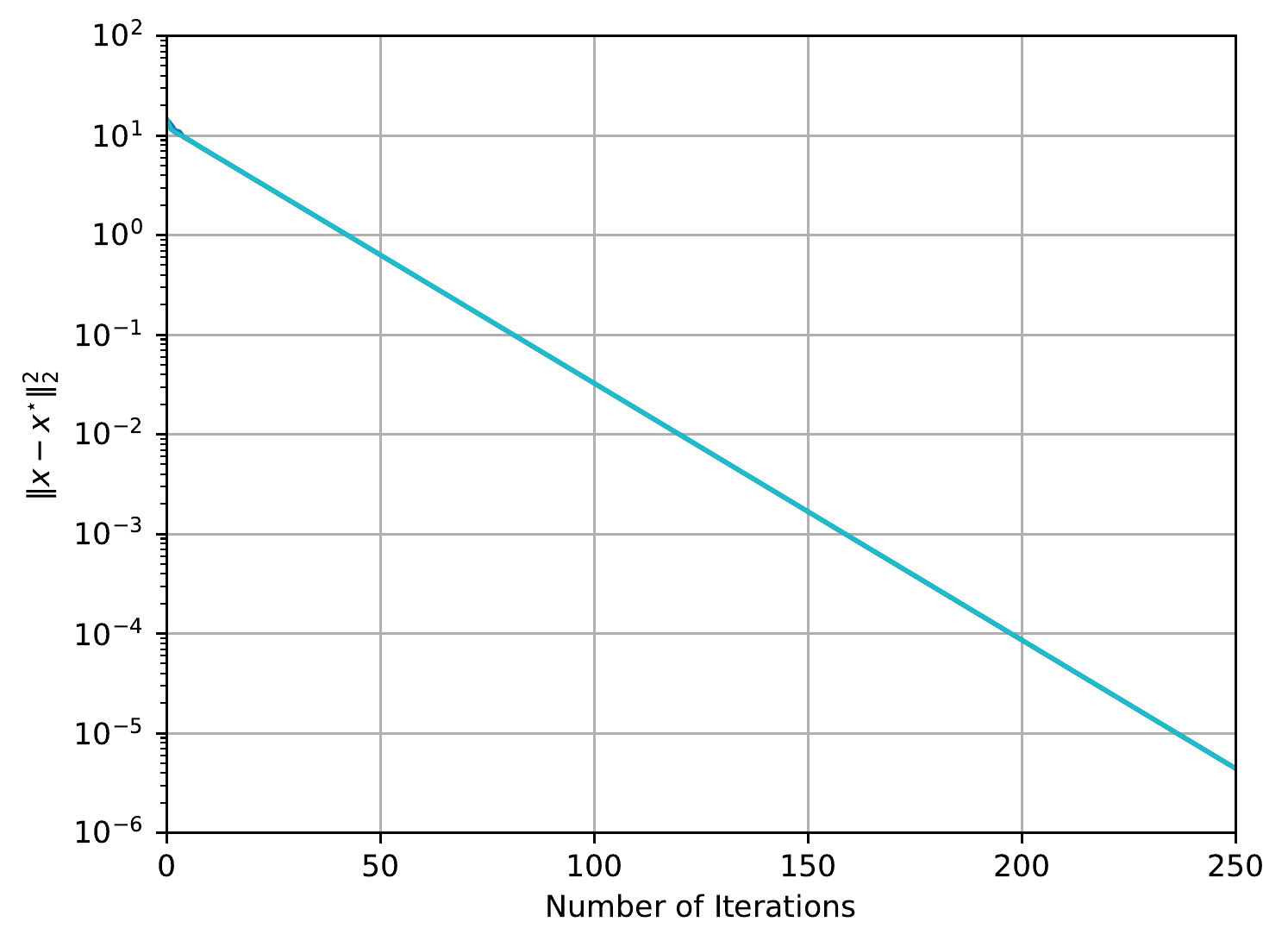}
        \caption{Convergence of the iterates computed by each robot using C-ADMM, implemented on hardware, on the optimization problem with three robots in \eqref{eqn:factored_dist_eqn}. The convergence errors of all the robots overlap in the figure.}
        \label{fig:hardware_convergence}
    \end{figure}	
    
    {As C-ADMM is robust to wide range of penalty parameters (as in Fig.~\ref{fig:BLE_step_sense}),} we set the penalty parameter in C-ADMM to a value of $5$ and do not perform a comprehensive search for the penalty parameter. In our experiments, this value of the penalty parameter provided suitable performance. In Figure \ref{fig:hardware_convergence}, we show the convergence error between the iterates of each robot and the global solution, which is obtained by aggregating the local data of all robots and then computing the solution centrally. The convergence errors of all the robots' iterates overlap in the figure, with the error decreasing below $10^{-5}$ within $250$ iterations, showing convergence of the local iterates of each robot to the optimal solution. { Again, due the severe bandwidth limitations of the XBee radios, these 250 iterations corresponded to approximately 23 mins of wall clock time, of which approximately 99.97\% was due to communication overhead.  With a well-engineered 5 Gbit/sec WiFi or 5G implementation, we expect this wall clock time for executing the 250 iterations of C-ADMM shown in Fig.~\ref{fig:hardware_convergence} to take  approximately $0.005$ sec.
    
    This small-scale experiment reveals several of the important considerations in implementing distributed optimization algorithms using physical communication hardware. First, while synchrony is crucial for certain methods including C-ADMM, we can satisfy this requirement even on relatively simple equipment by using a barrier strategy. Second, bandwidth limitations highlight the importance of considering low-dimensional representations of the state of the problem and/or quantization methods. For instance, communicating the optimization variable requires fewer broadcast rounds than communicating the measurements in the example problem that we considered. Finally, tuning is an important consideration, and C-ADMM provides a suitable solution due to its robustness to the choice of the $\rho$ parameter. }

\section{Conclusion}
\label{sec:conclusion}
In this tutorial, we have demonstrated that a number of canonical problems in multi-robot systems can be formulated and solved through the framework of distributed optimization. We have identified three broad classes of distributed optimization algorithms: distributed first-order methods, distributed sequential convex programming methods, and the alternating direction method of multipliers (ADMM). Further, we have described the optimization techniques employed by the algorithms within each category, providing a representative algorithm for each category. In addition, we have demonstrated the application of distributed optimization in simulation, on a distributed multi-drone vehicle tracking problem, and on hardware, showing the practical effectiveness of distributed optimization algorithms. However, important challenges remain in developing distributed algorithms for constrained, non-convex robotics problems, and algorithms tailored to the limited computation and communication resources of robot platforms, which we discuss in greater detail in the second paper in this series \cite{shorinwa_distributed_2023-1}.

\section*{Acknowledgment}
The authors would like to thank Siddharth Tanwar for implementing the C-ADMM multi-drone target tracking algorithm on XBee networking hardware.

	\bibliographystyle{IEEEtran}
	\bibliography{references}

\end{document}